\documentclass[lettersize,journal]{IEEEtran}
\usepackage{amsmath,amsfonts}
\usepackage{algorithmic}
\usepackage{algorithm}
\usepackage{array}
\usepackage[caption=false,font=normalsize,labelfont=sf,textfont=sf]{subfig}
\usepackage{textcomp}
\usepackage{stfloats}
\usepackage{url}
\usepackage{verbatim}
\usepackage{graphicx}
\usepackage{cite}
\usepackage{amssymb}
\usepackage{array}
\usepackage{mathrsfs}
\usepackage{stmaryrd}
\usepackage{hyperref}
\usepackage{booktabs}
\usepackage{multirow}
\usepackage{color}

\newcommand{\norm}[1]{\left\lVert \, #1 \, \right\rVert}

\newcommand{\KOp}[1]{\llbracket #1 \rrbracket} 

\newcommand{\vect}[1]{\mathbf{#1}}
\newcommand{\va}{\mathbf{a}}
\newcommand{\vb}{\mathbf{b}}
\newcommand{\vc}{\mathbf{c}}

\newcommand{\vx}{\mathbf{x}}

\newcommand{\vz}{\mathbf{z}}

\newcommand{\vmu}{\boldsymbol{\mu}}

\newcommand{\matr}[1]{\mathbf{#1}}
\newcommand{\mA}{\matr{A}}
\newcommand{\mB}{\matr{B}}
\newcommand{\mC}{\matr{C}}
\newcommand{\mD}{\matr{D}}
\newcommand{\mE}{\matr{E}}
\newcommand{\mF}{\matr{F}}
\newcommand{\mG}{\matr{G}}
\newcommand{\mH}{\matr{H}}
\newcommand{\mI}{\matr{I}}

\newcommand{\mX}{\matr{X}}
\newcommand{\mY}{\matr{Y}}
\newcommand{\mZ}{\matr{Z}}
\newcommand{\mM}{\matr{M}}

\newcommand{\mDelta}{\matr{\Delta}}

\newcommand{\mP}{\matr{P}}

\newcommand{\R}{\mathbb{R}}

\newcommand{\diag}[1]{\text{Diag}\left(#1\right)}   
\makeatletter
\newcommand{\argmin}{\mathop{\text{~argmin~}}}

\DeclareMathOperator{\prox}{prox}  
\DeclareMathOperator{\vecn}{vec}

\newcommand{\tens}[1]{\boldsymbol{\mathcal{#1}}}

\newcommand{\tT}{\tens{T}}
\newcommand{\tX}{\tens{X}}
\newcommand{\tY}{\tens{Y}}
\newcommand{\tZ}{\tens{Z}}

\newcommand{\tN}{\tens{N}}

\begin{document}

\title{PARAFAC2-based Coupled Matrix and Tensor Factorizations with Constraints}

\author{
Carla Schenker, \thanks{Carla Schenker is with Simula Metropolitan Center for Digital Engineering, Oslo, Norway. E-mail: carla@simula.no.}
Xiulin Wang, 
\thanks{Xiulin Wang is with the First Affiliated Hospital of Dalian Medical University, Dalian, China and Dalian Innovation Institute of Stem Cell and Precision Medicine, Dalian, China. E-mail: xiulin.wang@foxmail.com}
David Horner, \thanks{David Horner is with COPSAC: Copenhagen Studies on Asthma in Childhood, Herlev-Gentofte Hospital, Copenhagen, Denmark. E-mail: david.horner@dbac.dk}
Morten A. Rasmussen, \thanks{Morten A. Rasmussen is with the University of Copenhagen, Frederiksberg, Denmark and COPSAC: Copenhagen Studies on Asthma in Childhood, Herlev-Gentofte Hospital, Copenhagen, Denmark. E-mail: mortenr@food.ku.dk}
Evrim Acar\thanks{Evrim Acar is with Simula Metropolitan Center for Digital Engineering, Oslo, Norway. E-mail: evrim@simula.no.}
}

\markboth{Journal of \LaTeX\ Class Files,~Vol.~14, No.~8, August~2021}%
{Shell \MakeLowercase{\textit{et al.}}: A Sample Article Using IEEEtran.cls for IEEE Journals}


\maketitle

\begin{abstract}
Data fusion models based on Coupled Matrix and Tensor Factorizations (CMTF) have been effective tools for joint analysis of data from multiple sources. While the vast majority of CMTF models are based on the strictly multilinear CANDECOMP/PARAFAC (CP) tensor model, recently also the more flexible PARAFAC2 model has been integrated into CMTF models. PARAFAC2 tensor models can handle irregular/ragged tensors and have shown to be especially useful for modelling dynamic data with unaligned or irregular time profiles.
However, existing PARAFAC2-based CMTF models have limitations in terms of possible regularizations on the factors and/or types of coupling between datasets. To address these limitations, in this paper we introduce a flexible algorithmic framework that fits PARAFAC2-based CMTF models using Alternating Optimization (AO) and the Alternating Direction Method of Multipliers (ADMM). The proposed framework allows to impose various constraints on all modes and linear couplings to other matrix-, CP- or PARAFAC2-models.
Experiments on various simulated and real data demonstrate the utility and versatility of the proposed framework as well as its benefits in terms of accuracy and efficiency in comparison with state-of-the-art methods.

\end{abstract}


\section{Introduction}

\IEEEPARstart{D}{ata} about a phenomenon of interest can often be obtained from multiple sources containing complementary information. For instance, multiple neuroimaging modalities such as EEG (electroencephalography) and fMRI (functional Magnetic Resonance Imaging) signals with complementary spatial and temporal resolutions are often collected to better understand the functioning of the brain \cite{LaAdJu15, AcShLe19}. Similarly, to detect early life risk factors or early signs of various diseases, multi-omics datasets including genetic, longitudinal metabolomics and microbiome data are collected \cite{PrMaEa17,BeVaDr20}. Joint analysis of such datasets from multiple sources, also referred to as data fusion, has the potential to exploit these complementary measurements and enhance knowledge discovery \cite{LaAdJu15,AcBrSm15}. Such multimodal data are often heterogeneous consisting of datasets in the form of matrices and higher-order tensors (\textit{i.e.} multiway arrays with more than two axes of variation), and some change in time while some are static. For instance, longitudinal metabolomics measurements can be represented as a third-order tensor with modes: subjects, metabolites and time \cite{YaLiHo24}. When jointly analyzing such datasets with the goal of extracting insights, it is important to extract interpretable patterns and account for modality specific structure, \textit{e.g.} the right model accounting for time-evolving patterns or individual (subject-specific) differences.

Coupled matrix and tensor factorizations (CMTF) have been an effective approach for joint analysis of such datasets that can be represented in the form of matrices and higher-order tensors in many domains including social network analysis \cite{ErAcCe13,LiSuCaKo09}, bioinformatics \cite{PoSaVaAl11,Ba08}, neuroscience \cite{chatzichristos2022coupled,AcShLe19} and metabolomics \cite{AcBrSm15,AcGuRa12,LiYaHo24}. CMTF models use a low-rank model to approximate each dataset, where the factors/patterns of some modes are common or related across multiple datasets. These couplings between different datasets can be a hard coupling, \textit{i.e.} exact equality, or a soft/flexible coupling, \textit{i.e.} only approximately equal \cite{ChDaEsKoTh18,RiDuGu15}. Since datasets often have both shared, \textit{i.e.} common to several modalities, and unshared, \textit{i.e.} individual to one modality, components, models with partial couplings are frequently used \cite{AcPaGu14,ChDaEsKoTh18}. Couplings with (linear) transformations have also been effective in terms of handling different spatial, temporal, or spectral relationships between datasets \cite{FaCoCo16,AlKaSi20,KaFuSi18,RiDuGu15,EyHuDe17}. Furthermore, additional constraints and regularizations on the factor matrices such as non-negativity, sparsity or smoothness are often used to obtain more meaningful patterns and improve identifiability of CMTF models \cite{EyHuDe17,LiSuCaKo09,AcGuRa12,ApCuAg18}.

CMTF models typically use the CANDECOMP/PARAFAC (CP) model \cite{Ha70,CaCh70} to approximate higher-order data tensors. The CP decomposition models a tensor as the sum of rank-one tensors and thus assumes that the data follows a strictly multilinear structure. The PARAFAC2 model \cite{Hars1972b,KiTeBr99}, on the other hand, has more relaxed assumptions and allows one factor matrix to vary across tensor slices which makes it particularly useful to account for subject-specific variations in functional neuroimaging data analysis \cite{MaCh17}, to model temporal data with irregular or unaligned time profiles \cite{BrAnKi99,PePa19,AfPePa18} or evolving patterns \cite{RoBhJi20,acar2022tracing}. Recently, PARAFAC2 has been incorporated in CMTF models \cite{ChDaEsKoTh18, ChNaHa18, AfPePa20, GuThPa20, SoHiMa23}. For instance, a PARAFAC2-based CMTF model is used to fuse EEG and fMRI data accounting for subject variability in EEG through the use of PARAFAC2 \cite{ChDaEsKoTh18}, where soft (and partial) couplings are possible in the subject mode, as well as the temporal and spatial mode via linear transformations. Coupled PARAFAC2 models have also been proposed to fuse EEG and MEG (magnetoencephalogram) data sets \cite{ChNaHa18}. In temporal phenotyping, a PARAFAC2-based CMTF model called TASTE (Temporal and Static Tensor Factorization), has been used to jointly analyze EHR (electronic health record) data and patient demographic information \cite{AfPePa20}. The EHR data which consists of irregular and unaligned longitudinal clinical visits is modeled using  a non-negative PARAFAC2 model coupled with a non-negative matrix factorization that models the static patient demographics. The C$^3$APTION framework \cite{GuThPa20} extends TASTE to the coupling of a (non-negative) PARAFAC2 model with a (non-negative) CP model. Most recently, coupled PARAFAC2 models have also been used in chemometrics to analyze a $4th$- (or even higher-) order tensor which has two (or more) modes with drifts \cite{SoHiMa23}. Two different unfoldings into $3rd$-order tensors are modeled simultaneously using coupled PARAFAC2 decompositions to account for each of the two modes with retention time drifts. Despite the recent progress, available PARAFAC2-based CMTF models are limited in terms of constraints and regularizations that can be imposed on the factor matrices or types of couplings between datasets. In particular, they are all limited in the types of constraints that can be enforced on the varying mode of PARAFAC2. This mode is either estimated implicitly \cite{ChDaEsKoTh18,ChNaHa18}, making it difficult to impose any constraints, or the approach of a \textit{flexible} PARAFAC2 constraint from \cite{CoBr18} is adopted \cite{AfPePa20,GuThPa20,SoHiMa23}, which allows for non-negativity constraints.

In this paper, we propose a flexible, efficient and accurate algorithmic framework for PARAFAC2-based CMTF models that employs Alternating Optimization (AO) - Alternating Direction Method of Multipliers (ADMM). The framework is flexible in the sense that many types of constraints and regularizations as well as linear couplings in the static modes of PARAFAC2 with either a matrix-, a CP- or another PARAFAC2-decomposition can be imposed in a plug-and-play fashion. By using ADMM also for the PARAFAC2 constraint, we enable the use of various constraints even for the varying PARAFAC2 mode.
We demonstrate that our proposed method has at least comparable and often better performance in terms of accuracy and computation time compared to state-of-the-art methods for non-negativity constraints and exact coupling on synthetic datasets in a number of different settings. Especially, it consistently achieves better PARAFAC2 structure. We also demonstrate that the proposed approach leads to accurate solutions for linear couplings and other constraints than non-negativity. Furthermore, previous studies using PARAFAC2-based CMTF models use PARAFAC2 to cope with individual or unaligned time profiles. However, to the best of our knowledge, CMTF models with PARAFAC2 have never been used to model dynamic data with evolving components, similar to the work in \cite{RoBhJi20}, together with static side-information. Here, we demonstrate the promise of the proposed model in terms of jointly analyzing static and dynamic data with evolving components on a synthetic dataset and in a real metabolomics application.

This paper is an extension of our preliminary results \cite{ScWaAc23} and includes a more detailed description of the algorithm and introduces couplings in the mode across which the variation in PARAFAC2 occurs, commonly denoted as mode $\mC$. Couplings in this mode are more complicated than couplings in the static mode of PARAFAC2, commonly denoted as mode $\mA$, which were studied in \cite{ScWaAc23}. Moreover, this work includes extensive comparisons with state-of-the-art methods TASTE and C$^3$APTION, and a novel application in metabolomics.


    

After establishing our notation, in Section \ref{sec:CMTF models} we introduce the type of general PARAFAC2-based CMTF models that our proposed framework can fit. Section \ref{sec:other algos} gives an overview of state-of-the-art methods and in Section \ref{sec:AO-ADMM} we present the proposed AO-ADMM framework. Finally, numerical experiments on synthetic as well as a real dataset are presented in Section\ref{sec:experiments}.

\subsection{Notation}
We denote tensors by boldface uppercase calligraphic letters $\tT$, matrices by boldface uppercase letters $\mM$, vectors by boldface lowercase letters $\vect{v}$ and scalars by lowercase letters $a$.
$\norm{\cdot}_F$ denotes the Frobenius norm. The characteristic function of a set $\mathcal{D}$ is given by $\iota_{\mathcal{D}}$. $\ast$ denotes the (element-wise) Hadamard product between two equally sized matrices, $\otimes$ denotes the Kronecker product of two matrices 
and $\odot$ denotes the Khatri-Rao product. For more details, see \cite{KoBa09}.

\section{PARAFAC2-based CMTF models}\label{sec:CMTF models}
Before we present PARAFAC2-based CMTF models, we give a short introduction to CP and PARAFAC2 decompositions. After that, we will first formally introduce PARAFAC2-based CMTF models which are coupled in mode $\mA$, followed by models coupled in mode $\mC$.

\subsection{CP and PARAFAC2}
\subsubsection{CP decomposition}
The Canonical Polyadic Decomposition (CP/CPD), also called  CANDECOMP (canonical decomposition) or PARAFAC (parallel factors) \cite{Hi27a,CaCh70,Ha70}, represents a tensor as a finite sum of rank-one tensors. For a $3$-way tensor $\tX \in \R^{I \times J \times K}$, the $R$-component CP decomposition is defined as
\begin{equation}\label{eq:CPdef}
  \tX \approx \sum_{r=1}^R \va_r \circ \vb_r \circ \vc_r:= \KOp{\mA,\mB,\mC },  
\end{equation}
where the vectors $\va_r \in \R^{I}$, $\vb_r \in \R^{J}$ and $\vc_r \in \R^{K}$ are called factor vectors and together they form the $r$-th component of the CP decomposition. The factor vectors are the columns of factor matrices $\mA \in \R^{I \times R},\mB \in \R^{J \times R}$ and $\mC \in \R^{K \times R}$, respectively. The (CP-) rank of a tensor $\tX$ is defined as the minimum number of rank-one tensors that generate $\tX$ as their sum \cite{Hi27a}. Thus, when $R$ is equal to the rank of $\tX$, then \eqref{eq:CPdef} holds with exact equality. When analyzing real-world datasets, however, $R$ is typically chosen much smaller than the true rank, and the CP decomposition \eqref{eq:CPdef} then constitutes a (low-rank) approximation of $\tX$. The (full-rank) CP decomposition is unique (up to scaling and permutation ambiguities) under mild conditions \cite{Kruskal77,SiBr00,DoLa13b,EvLa22}.
An alternative representation of the CP decomposition in terms of the frontal slices of $\tX$, $\mX_k \in \R^{I\times J}$, $k\leq K$, is the following,
\begin{equation*}
    \mX_k \approx  \mA \mD_k \mB^T, \ \ \  \text{for} \ \  k=1,...,K,
\end{equation*}
where the diagonal matrix $\mD_k \in \R^{R \times R}$ contains the $k$th row of $\mC$ on its diagonal, \textit{i.e.} $\mD_k=\text{Diag}(\mC(k,:))$. This means that each slice $\mX_k$ is modelled by the same patterns in $\mA$ and $\mB$ and those patterns can only change in $\textit{strength}$, \textit{i.e.} a scalar weight given by the values on the diagonal of $\mD_k$, across the slices $k=1,...,K$.

\subsubsection{PARAFAC2}\label{sec:PARAFAC2}
The PARAFAC2 model relaxes this strict multilinearity assumptions of the CP decomposition and allows one mode of the decomposition to vary across another mode. This also makes the decomposition of so-called \textit{ragged} tensors with slices of different lengths possible. PARAFAC2 approximates the slices of a (ragged) $3$-way tensor $\tX$, $\mX_k \in \R^{I\times J_k}$, $k\leq K$, with a (low-rank) factorization of rank $R$ as follows \cite{Hars1972b},
 \begin{equation*}
 \begin{aligned}
\mX_k &\approx  \mA \mD_k \mB_k^T, \ \ \ \mB_k \in \mathcal{P},  \ \ \ \text{for} \ \  k=1,...,K,\\
& \mathcal{P} = \left\lbrace\left\lbrace \mB_k \right\rbrace_{k=1}^K \mid \mB_{k}^T\mB_{k}=\mathbf{\Phi},\ k=1,..., K \right\rbrace,
 \end{aligned}
\end{equation*}
with factor matrices  $\mA \in \R^{I \times R}$ and $\mB_k \in \R^{J_k \times R}$ and a diagonal matrix $\mD_k \in \R^{R \times R}$. With factor matrix $\mC\in\mathbb{R}^{K\times R}$, we denote the matrix that contains the diagonals of $\mD_k$ as rows, $\mD_k=\text{Diag}(\mC(k,:))$, as before. Thus, unlike the CP model, 
PARAFAC2 allows for the patterns in $\mB$ to vary across the slices $k$. However, in order to ensure a unique decomposition, the cross product of the factor matrices $\mB_k$ should be invariant w.r.t. $k$. This is called the PARAFAC2 constraint, denoted as $\mB_k \in \mathcal{P}$. It implies that for centered scores $\mB_k$, the correlations between the components are constant over $k$ since scale differences are captured in the $\mD_k$ matrices \cite{KiTeBr99}. 
An equivalent formulation of the PARAFAC2 constraint is to require that each matrix $\mB_k$ can be expressed as the product of an orthogonal matrix $\mP_k$ and an invariant matrix $\mDelta_B$, as follows
{\small\begin{equation}\label{eq:PAR2constralt}
\mB_k \in \mathcal{P} = \left\lbrace\left\lbrace \mB_k \right\rbrace_{k=1}^K \mid, \mB_k = \mP_k \mDelta_B, \mP_k^T \mP_k= \mI\ k=1,..., K \right\rbrace.
\end{equation}}
The PARAFAC2-ALS algorithm \cite{KiTeBr99}, which is the standard algorithm for fitting PARAFAC2 decompositions, makes use of this reformulation. It solves the following optimization problem, where $\mB_k$ has been substituted by $\mP_k \mDelta_B$:
\begin{equation}\label{eq:PAR2ALSobjectivetrans}
\begin{aligned}
    & \underset{
       \mA,\mDelta_B ,\left\lbrace\mD_k, \mP_k\right\rbrace_{k\leq K}}{\argmin} & &
  \sum\limits_{k=1}^{K} \norm{\mX_k - \mA \mD_k \mDelta_B^T \mP_k^T}_F^2  \\
   & \ \ \ \ \ \ \ \ \text{s.t. } & &
    \mP_k^T \mP_k = \mI , \ \ \ k=1,...,K
\end{aligned}
\end{equation}
First, PARAFAC2-ALS solves for $\mP_k$, keeping all other variables fixed, which means solving an orthogonal Procrustes problem \cite{GoLo13}. Then, for fixed $\mP_k$, since $\mP_k$ is orthogonal, problem \eqref{eq:PAR2ALSobjectivetrans} is equivalent to
\begin{equation*}
    \underset{
       \mA,\mDelta_B ,\left\lbrace\mD_k \right\rbrace_{k\leq K}}{\argmin} 
  \sum\limits_{k=1}^{K} \norm{\mX_k\mP_k - \mA \mD_k \mDelta_B^T }_F^2.  
\end{equation*}
This is exactly the problem of fitting a CP model to the tensor $\tY$ with frontal slices $\mY_k = \mX_k\mP_k$, which is done using standard ALS for CP decompositions \cite{CaCh70,Ha70}.
It has been shown that the PARAFAC2 decomposition is unique (again up to scaling and permutation ambiguities) under certain conditions \cite{HaLu96,BeJoKi96,KiTeBr99}.
\subsection{PARAFAC2-based CMTF models coupled in mode $\mA$}
\begin{figure}[t!]
\centering
\includegraphics[width=\columnwidth]{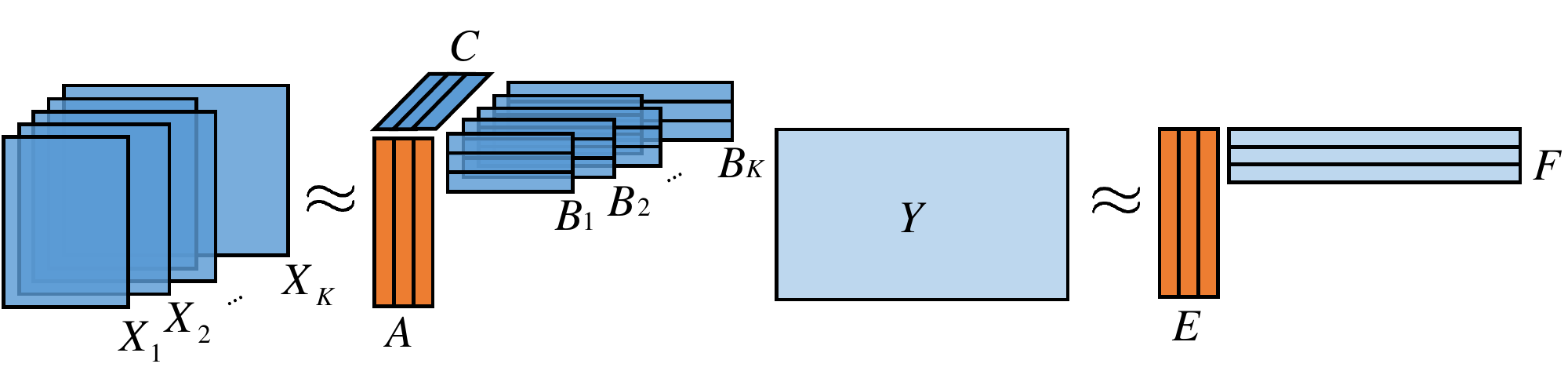}
\caption{A PARAFAC2 model coupled with a matrix factorization.}
\label{fig:syn1B}
\end{figure}

For the sake of readability, we present here a PARAFAC2-based CMTF model where a PARAFAC2 decomposition of a third-order tensor is linearly coupled in the first mode, mode $\mA$, with a matrix factorization (as in Figure \ref{fig:syn1B}). Note that it is in general also possible to simultaneously decompose more than two datasets by imposing several couplings on different modes and using either PARAFAC2, CP or matrix decompositions. An example of a model with PARAFAC2 coupled in the third mode, mode $\mC$, with a CP decomposition is presented in Section \ref{sec:modelinC}. 

Here, we assume that the slices of a (ragged) tensor $\tX$, $\mX_k \in \R^{I_1\times J_k}$, $k\leq K$, can be approximated with a PARAFAC2 decomposition of rank $R_1$ as 
 \begin{equation*}
 \small
\mX_k \approx \mA \mD_k \mB_k^T, \ \ \ \mB_k \in \mathcal{P},  \ \ \ \text{for} \ \  k=1,...,K,
\end{equation*}
with factor matrices $\mA \in \R^{I_1 \times R_1}$, $\mB_k \in \R^{J_k \times R_1}$ and $\mD_k \in \R^{R_1 \times R_1}$.
We also assume that the data matrix ${\mY}\in\mathbb{R}^{I_2\times L}$ can be factorized using $R_2$ components as ${\mY} \approx {\mE \mF}^{T}$, with ${\mE}\in\mathbb{R}^{I_2\times R_2}$ and ${\mF}\in\mathbb{R}^{L\times R_2}$, and that the factor matrices of the first mode from both datasets, $\mA$ and $\mE$, are linearly coupled.
This means that the coupling can be written as 
\begin{equation*}
    \begin{aligned}
       \mH_{\mA} \vecn(\mA) &= \mH_{\mA}^{\Delta} \vecn(\mDelta), \\
       \mH_{\mE} \vecn(\mE) &= \mH_{\mE}^{\Delta} \vecn(\mDelta),
    \end{aligned}
\end{equation*}
with some unknown generating variable $\mDelta\in\mathbb{R}^{m_1 \times m_2}$ and known transformation matrices $\mH_{\mA} \in \R^{h \times R_1 I_1}, \mH_{\mE} \in \R^{h \times R_2 I_2}, \mH_{\mA}^{\Delta} \in \R^{h \times m_1 m_2}, \mH_{\mE}^{\Delta} \in \R^{h \times m_1 m_2}$ as in \cite{ScCoAc21}. 
These transformations can be used to model different linear relationships between datasets as, for instance, averaging, blurring and downsampling \cite{KaFuSi18}, convolutions \cite{ChDaEsKoTh18}, and also partially shared components. For more details and examples about these types of linear couplings, we refer to \cite{ScCoAc21}.
We furthermore allow each mode of the coupled model to be regularized by a regularization function $g$. We allow any proper closed convex function $g$ as regularization, which ensures that the proximal operator of $g$ exists and is unique.
Hard constraints are also included via characteristic functions of convex sets, \textit{e.g.,} $g=\iota_{\R^+}$ for non-negativity. Other examples of important regularizations and constraints that can be imposed are sparsity, ridge regularization, smoothness, normalization, monotonicity or simplex constraints, see also \cite{ScCoAc21,RoScCa22} for more details. In practice, also non-convex functions can be used as long as their proximal operator is computable, but their proximal operator may not be unique any more.
Using the squared Frobenius norm loss and scalar weights $w_1, w_2$, the overall coupled factorization problem is then formulated as follows:
{\small\begin{align}
    & \underset{\underset{
       \mA,\mE,\mF,\mDelta}{\left\lbrace\mD_k, \mB_k\right\rbrace_{k}}}{\argmin} & & 
   \sum\limits_{k=1}^{K}\left[w_1  \norm{\mX_k  -  \mA \mD_k \mB_k^T }_F^2  +  g_D( \mD_k )  +  g_B(\mB_k)\right] \nonumber \\
   & & & \  +  w_2 \norm{\mY  -  \mE \mF^T}_F^2  +  g_A(\mA)  +  g_E(\mE)  +  g_F(\mF) \nonumber \\
   & \ \ \ \ \ \ \ \ \text{s.t. } & & \left\lbrace\mB_k\right\rbrace_{k\leq K} \in \mathcal{P},\label{eq:costfuncA} \\
   & & & \mH_{ \mA}  \vecn( \mA )  =  \mH_{ \mA}^{ \Delta}  \vecn( \mDelta ) ,\ \mH_{\mE}  \vecn(\mE)  =  \mH_{ \mE}^{ \Delta}  \vecn( \mDelta ).\nonumber
\end{align}}
This model is illustrated in Figure~\ref{fig:syn1B}.

\begin{figure}[t!]
\centering
\includegraphics[width=\columnwidth]{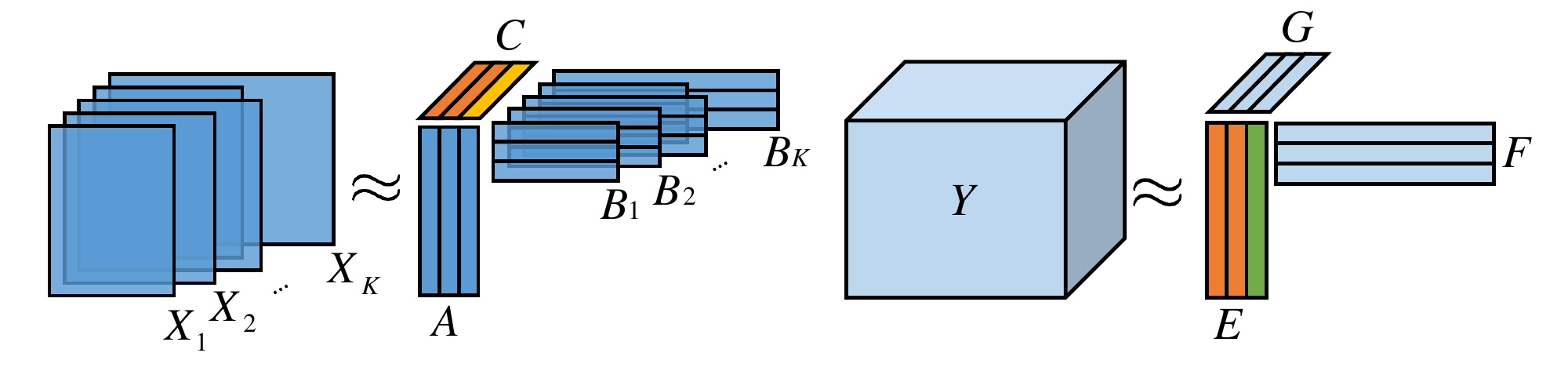}
\caption{A PARAFAC2 model partially coupled with a CP model.}
\label{fig:syn3B}
\end{figure}
\subsection{PARAFAC2-based CMTF models coupled in mode $\mC$}\label{sec:modelinC}
Here, we present a PARAFAC2-based CMTF model, where a PARAFAC2 model is linearly coupled with a CP decomposition in the third mode $\mC$ of PARAFAC2 (as in Figure \ref{fig:syn3B}). 
We again assume that the slices of a (ragged) tensor $\tX$, $\mX_k \in \R^{I_1\times J_k}$, $k\leq K$, can be approximated with a PARAFAC2 decomposition of rank $R_1$ as before, and the third-order data tensor $\tY \in \mathbb{R}^{I_2\times L \times P}$ can be approximated by a CP decomposition of rank $R_2$ as ${\mY} \approx \KOp{\mE, \mF,\mG}$, with factor matrices ${\mE}\in\mathbb{R}^{I_2\times R_2}$, ${\mF}\in\mathbb{R}^{L\times R_2}$ and ${\mG}\in\mathbb{R}^{P\times R_2}$. In this model, the factor matrices of modes $\mC$ and $\mE$ are assumed to be linearly coupled.
Again, all modes can also be regularized via proper closed convex functions $g$. The model is then given by
{\small
\begin{align}
    & \underset{\underset{
       \mA,\mE,\mF,\mG,\mDelta}{\left\lbrace \mB_k\right\rbrace_{k},\mC}}{\argmin} & & 
  \sum\limits_{k=1}^{K}\left[w_1 \norm{\mX_k - \mA \diag{\mC(k,:)} \mB_k^T}_F^2  + g_B(\mB_k)\right]\nonumber \\
   & & & \ + w_2 \norm{\tY - \KOp{\mE, \mF,\mG}}_F^2 + g_A(\mA) + g_E(\mE) \nonumber  \\
   & & & \  + g_C(\mC) + g_F(\mF) + g_G(\mG)\nonumber\\
   & \ \ \ \ \ \text{s.t. } & & \left\lbrace\mB_k\right\rbrace_{k\leq K} \in \mathcal{P},\label{eq:costfuncC}\\
   & & & \mH_{\mC} \vecn(\mC) = \mH_{\mC}^{\Delta} \vecn(\mDelta),\mH_{\mE} \vecn(\mE) = \mH_{\mE}^{\Delta} \vecn(\mDelta).\nonumber
\end{align}}
Note that in contrast to model \eqref{eq:costfuncA}, here we state the model in terms of $\mC$ and not $\mD_k$ for better readability.
This model with partially shared components is illustrated in Figure~\ref{fig:syn3B}.
\section{Algorithms for PARAFAC2-based CMTF models}\label{sec:other algos}
Previous studies that incorporate a PARAFAC2 decomposition in a CMTF model rely on algorithmic approaches either implicitly handling the PARAFAC2 constraint through the reformulation of the constraint or using a \textit{flexible} PARAFAC2 constraint \cite{CoBr18}. Both previously mentioned neuroimaging applications \cite{ChDaEsKoTh18,ChNaHa18} use the reformulation \eqref{eq:PAR2ALSobjectivetrans} of the PARAFAC2 constraint. In \cite{ChNaHa18}, the model is then fitted using ALS and in \cite{ChDaEsKoTh18}, it is fitted as a coupled matrix decomposition model within the Structured Data Fusion (SDF) framework \cite{SoBaLa15} of Tensorlab \cite{Vervliet2016Tensorlab}. However, since factor matrices $\mB_k$ of the varying mode of PARAFAC2 are estimated implicitly in both cases, it is not possible to impose constraints on them. Using the SDF framework, PARAFAC2-based CMTF models with partial and other linear couplings can be solved as well as models with various constraints on the \textit{static} factor matrices, as long as the constraints are supported by Tensorlab. In spite of that, Tensorlab does not currently explicitly support PARAFAC2 models. The TASTE \cite{AfPePa20} and C$^3$APTION frameworks \cite{GuThPa20}, on the other hand, adapt the idea of the flexible PARAFAC2 constraint \cite{CoBr18} to impose constraints also on the varying mode $\mB$. TASTE and C$^3$APTION are described in more detail here since we compare them to our proposed approach in the experiments.
\subsection{The TASTE framework}\label{sec:TASTE}
In TASTE \cite{AfPePa20}, a non-negative PARAFAC2 model is coupled in the $\mC$-mode with a non-negative matrix factorization. Using the idea of the flexible PARAFAC2 constraint from \cite{CoBr18}, the PARAFAC2 constraint is imposed as a regularization term which allows for non-negativity constraints on factor matrices $\mB_k$.
In the notation used here, the following model is solved:
{\small
\begin{align}
    & \underset{\underset{
       \mA,\mF,\mDelta_B}{\left\lbrace \mB_k,\mP_k\right\rbrace_{k},\mC}}{\argmin} & & 
  \sum\limits_{k=1}^{K}\left[\frac{1}{2} \norm{\mX_k - \mA \diag{\mC(k,:)} \mB_k^T}_F^2  + \iota_{\R^+}(\mB_k)\right]\nonumber \\
   & & & \ + \frac{\lambda}{2} \norm{\mY - \mC \mF^T}_F^2 + \iota_{\R^+}(\mA) + \iota_{\R^+}(\mC)+\iota_{\R^+}(\mF) \nonumber \\
   & & & \ +\sum\limits_{k=1}^{K}\left[\frac{\mu_k}{2} \norm{\mB_k - \mP_k \mDelta_B}_F^2\right]\nonumber \\
   & \ \ \ \ \ \text{s.t. } & & \mP_k^T \mP_k = \mI,\label{eq:costfuncTASTE}
\end{align}}
The model is fitted with an alternating optimization scheme, solving for one matrix at a time. The subproblems for matrices $\mP_k$, $k=1,...,K$, are Orthogonal Procrustes problems, which can be solved exactly via the singular value decomposition \cite{GoLo13}. The subproblem for matrix $\mDelta_B$ is an unconstrained least-squares problem which reduces to the computation of a mean. The subproblems for all other matrices are non-negative least-squares (NNLS) problems which are solved using block principal pivoting (BPP) \cite{KiPa11}.

\subsection{The C$^3$APTION framework}\label{sec:CAPTION}
The C$^3$APTION framework \cite{GuThPa20} is essentially an extension of the TASTE framework to the coupling of a non-negative PARAFAC2 model with a non-negative CP model. It makes use of the same idea and imposes the PARAFAC2 constraint as a regularization. In the notation used here, the model solved in C$^3$APTION is the following:
{\small
\begin{align}
    & \underset{\underset{
       \mA,\mF,\mG,\mDelta_B}{\left\lbrace \mB_k,\mP_k\right\rbrace_{k},\mC}}{\argmin} & & 
  \sum\limits_{k=1}^{K}\left[\frac{1}{2} \norm{\mX_k - \mA \diag{\mC(k,:)} \mB_k^T}_F^2  + \iota_{\R^+}(\mB_k)\right]\nonumber \\
   & & & \ + \frac{\lambda}{2} \norm{\tY - \KOp{\mC, \mF,\mG}}_F^2 + \iota_{\R^+}(\mA) + \iota_{\R^+}(\mC) \nonumber \\
   & & &+\iota_{\R^+}(\mF)+\iota_{\R^+}(\mG) \nonumber \\
   & & & \ +\sum\limits_{k=1}^{K}\left[\frac{\mu_k}{2} \norm{\mB_k - \mP_k \mDelta_B}_F^2\right]\nonumber \\
   & \ \ \ \ \ \text{s.t. } & & \mP_k^T \mP_k = \mI,\label{eq:costfuncCAPTION}
\end{align}}
The C$^3$APTION framework offers three different possibilities to fit this model. The first possibility is equivalent to the algorithm proposed in TASTE, using BPP to solve the NNLS subproblems, while the second possibility is to use active set (ASET) \cite{HyPa08} instead of BPP.
The third proposed possibility does not actually solve the model \eqref{eq:costfuncCAPTION} with the PARAFAC2 constraint as a penalty term. Instead, it imposes exact PARAFAC2 structure by employing a version of the standard ALS algorithm for PARAFAC2, but using ALS with thresholding at zero to obtain non-negative factor matrices. It should be noted, however, that the last option, ALS with zero-thresholding, despite being fast, does not yield a good approximation to the solution of the non-negative least-squares problems in general \cite{KiHePa14}. \\
 
TASTE and C$^3$APTION frameworks do not support any linear couplings and they have only non-negativity constraints implemented. In theory, some other constraints could be included by employing specialized constrained least-squares solvers instead of NNLS solvers. Furthermore, they suffer, as described in the original flexible PARAFAC2 approach \cite{CoBr18}, from time-consuming tuning of the penalty parameter $\mu_k$. They may even perform sub-optimal since the heuristic scheme for increasing the penalty parameter after each iteration as proposed in \cite{CoBr18}, is not implemented in the codes accompanying TASTE and C$^3$APTION papers \cite{AfPePa20,GuThPa20}, but instead the constant penalty parameter $\mu_k=1$ is used.\\

It is worth noting that the algorithm for the coupled PARAFAC2-PARAFAC2 models proposed in  \cite{SoHiMa23} adopts a method similar to TASTE and C$^3$APTION frameworks. This algorithm leverages the same flexible coupling approach to impose constraints such as nonnegativity or unimodality on the varying modes of the PARAFAC2 models. The factor matrices are updated in an alternating manner, and (constrained) least-squares solvers are employed for the subproblems. The only distinction lies in the decompositions being only approximately coupled through a penalty term on the difference between the coupled factor matrices.

\section{AO-ADMM algorithm for PARAFAC2-based CMTF models}\label{sec:AO-ADMM}
Similar to TASTE and C$^3$APTION frameworks, we employ alternating optimization (AO) over the different modes. To be more specific, we alternately update all factor matrices of a single mode (across tensors), while the factor matrices of all other modes remain constant. This means that coupled factor matrices such as $\mA$ and $\mE$ in \eqref{eq:costfuncA} are updated jointly. Moreover, updates for unrelated factor matrices from distinct tensors can be computed in parallel as they do not depend on each other.

For each subproblem, we use ADMM, which is a primal-dual algorithm
for convex constrained optimization problems of the form
\begin{equation*}\begin{aligned}
        &\argmin_{\vx,\vz} &f(\vx) + g(\vz),\\
        & \ \ \text{ s.t. } &\mA \vx +\mB \vz = \vc\end{aligned}
\end{equation*}
where $f$ and $g$ are proper closed convex functions~\cite{Boyd2011Distributed}. ADMM makes use of the duality theory for convex optimization by alternatingly minimizing the augmented Lagrangian, here in scaled form~\cite{Boyd2011Distributed},
\begin{equation*}
        L(\vx,\vz,\vmu) = f(\vx) + g(\vz) + \frac{\rho}{2}\norm{\mA \vx + \mB \vz - \vc + \vmu}_2^2,
\end{equation*}
for a constant dual variable $\vmu$ and
maximizing the dual function $G(\vmu)=\min_{\vx,\vz}L(\vx,\vz,\vmu)$ w.r.t.
$\vmu$. This is done via a gradient ascent step. ADMM in its simplest form is summarized in Algorithm~\ref{alg:admmplain}.
\begin{algorithm}[!t]
        \begin{algorithmic}
            \WHILE{convergence criterion is not met}
            \STATE{$\small{\vx^{(k+1)}=\underset{\vx}{\argmin} f(\vx) + \frac{\rho}{2} \norm{\smash{ \mA\vx +
                    \mB\vz^{(k)} - \vc + \vmu^{(k)}}}_2^2}$}
            \STATE{$\small{\vz^{(k+1)}=\underset{\vz}{\argmin} g(\vz) +~\frac{\rho}{2} \norm{\smash{ \mA\vx^{(k+1)} +
                    \mB\vz - \vc + \vmu^{(k)}} }_2^2}$}
                    \STATE{$\small{\vmu^{(k+1)} = \vmu^{(k)} + \mA\vx^{(k+1)} + \mB\vz^{(k+1)} -\vc }$}
            \STATE{$\small{k = k+1}$}
            \ENDWHILE{}
        \end{algorithmic}
        \caption{Skeleton of scaled-form ADMM~\cite{Boyd2011Distributed}}\label{alg:admmplain}
\end{algorithm}
Based on primal and dual feasibility conditions, the stopping conditions proposed in \cite{Boyd2011Distributed} are that primal ($\vect{r}^{(k)}$) and dual residuals ($\vect{s}^{(k)}$) must be small, \textit{i.e.,}
\begin{equation}\label{eq:ADMMstopcrit}\begin{aligned}
\norm{\smash{\vect{r}^{(k)}}}_2 &= \norm{\smash{\mA\vx^{(k)}+\mB\vz^{(k)}-\vc}}_2\leq \epsilon^{\text{pri}},\\
\norm{\smash{\vect{s}^{(k)}}}_2 &= \norm{\smash{\rho \mA^T \mB \left(\vz^{(k+1)}-\vz^{(k)} \right)}}_2\leq \epsilon^{\text{dual}},
\end{aligned}
\end{equation}
where $\epsilon^{\text{pri}}>0$ and   $\epsilon^{\text{dual}}>0$ are feasibility tolerances, which can be set as follows:
\begin{equation*}\label{eq:ADMMstoptol}\small{\begin{aligned}
\epsilon^{\text{pri}}&=\sqrt{\text{length}(\vc)}\epsilon^{\text{abs}}+\epsilon^{\text{rel}}\max \left\lbrace \norm{\smash{\mA\vx^{(k)}}}_2,\norm{\smash{\mB\vz^{(k)}}}_2,\norm{\smash{\vc}}_2 \right\rbrace,\\
\epsilon^{\text{dual}}&=\sqrt{\text{length}(\vx)}\epsilon^{\text{abs}}+\epsilon^{\text{rel}}\rho \norm{\smash{\mA^T \vmu^{(k)}}}_2,
\end{aligned}}
\end{equation*}
with  $\epsilon^{\text{abs}}>0$ and $\epsilon^{\text{rel}}>0$ denoting the absolute and relative tolerance, respectively.

The subproblems for each constrained factor matrix (or coupled factor matrices) are transformed into ADMM form by introducing \textit{split} variables $\mZ$ which separate the factorization from the constraint, as first proposed for constrained CP decompositions in \cite{HuSiLi16}. For instance, for uncoupled factor matrix $\mA$ in model \eqref{eq:costfuncC}, the problem takes the form
{\small
\begin{align}
    & \underset{\mA,\mZ_A}{\argmin} & & 
  \sum\limits_{k=1}^{K}\left[w_1 \norm{\mX_k - \mA \diag{\mC(k,:)} \mB_k^T}_F^2  + g_A(\mZ_A)\right]\nonumber \\
   & \ \ \ \ \ \text{s.t. } & & \mA = \mZ_A.\nonumber
\end{align}}
Under the assumption that the regularization function $g_A$ is proper closed convex, ADMM can directly be applied to this problem, resulting in simple and efficient individual ADMM updates for $\mA$ and $\mZ_A$, respectively. Specifically, the update for $\mZ_A$ reduces computing the proximal operator corresponding to the function $g_A$. 
The proximal operator of a proper lower semi-continuous function $g$ is defined as\cite{Moreau1965Proximite,Be17}
\begin{equation*}
        \prox_{\rho g}(\vx) = \underset{\vect{u}}{\argmin} g(\vect{u}) + \frac{1}{2\rho} \norm{ \vx - \vect{u}
        }_2^2,
\end{equation*}
for any $\rho>0$, and it is single-valued for convex $g$.
Closed-form expressions are available for many commonly used regularization functions $g$. For others, efficient estimation algorithms exist, as shown in \cite{PaBo14,Be17} and here \footnote{\url{http://proximity-operator.net/proximityoperator.html}}.
This use of proximal operators facilitates the flexibility of AO-ADMM, as it allows to apply constraints and regularizations in a \textit{mix-and-match, plug-and-play} fashion \cite{HuSiLi16}.

There is an exception where ADMM cannot be directly applied. This exception is the subproblem for factor matrices $\mB_k$ of the varying mode of PARAFAC2 as the set $\mathcal{P}$ defined by the PARAFAC2 constraint is not convex.
We solve this, as proposed in our previous work \cite{RoScCa22}, by introducing two sets of \textit{split} variables, one for the PARAFAC2 constraint and one for the regularization given through $g_B$. The projection onto the non-convex set $\mathcal{P}$ is then efficiently approximated by an AO scheme. Note that this approach treats the PARAFAC2 constraint as a hard constraint, in contrast to TASTE and C$^3$APTION, where the PARAFAC2 constraint is added as a regularization term. Therefore, our approach does not face the problem of tuning a hyperparameter for the regularization strength.
We have already presented AO-ADMM for the PARAFAC2-based CMTF model \eqref{eq:costfuncA} with couplings in mode $\mA$ in \cite{ScWaAc23}.
Therefore, in the following, we only derive the ADMM algorithm for the subproblem of the third factor matrix $\mC$ of PARAFAC2 in the case that it is coupled. The updates are different from the ones previously presented for coupled mode $\mA$, since the factor matrices $\mB_k$ are allowed to vary across mode $\mC$ in the PARAFAC2 model. Following our previous work \cite{ScCoAc21,ScWaAc23}, we again consider five special types of linear couplings. 
ADMM updates for other modes of models \eqref{eq:costfuncA} and \eqref{eq:costfuncC} can be found in \cite{RoScCa22,ScWaAc23} and are also given in the supplementary material \footnote{\url{https://github.com/AOADMM-DataFusionFramework/Supplementary-Materials}} for convenience. 
The code is publicly available as part of a general AO-ADMM framework that can handle any number of coupled decompositions, including CP, PARAFAC2 and matrix decompositions \footnote{\url{https://github.com/AOADMM-DataFusionFramework}}.
\subsection{Algorithm for couplings in mode $\mC$}
For the sake of simplicity, we derive the updates for model \eqref{eq:costfuncC}, where a PARAFAC2 and a CP decomposition are linearly coupled in modes $\mC$ of PARAFAC2 and $\mE$ of the CP decomposition, \textit{i.e.} we want to solve the following subproblem for coupled factor matrices $\mC$ and $\mE$:
{\small
\begin{align}
    & \underset{
       \mC,\mE,\mDelta}{\argmin} & & 
  \sum\limits_{k=1}^{K}\left[w_1 \norm{\mX_k - \mA \diag{\mC(k,:)} \mB_k^T}_F^2 \right]\nonumber \\
   & & & \ + w_2 \norm{\tY - \KOp{\mE, \mF,\mG}}_F^2+ g_C(\mC) + g_E(\mE) \nonumber  \\
   & \ \ \ \ \ \text{s.t. } & & \mH_{\mC} \vecn(\mC) = \mH_{\mC}^{\Delta} \vecn(\mDelta),\mH_{\mE} \vecn(\mE) = \mH_{\mE}^{\Delta} \vecn(\mDelta)\nonumber
\end{align}}
We transform this problem into the following form,
{\small
\begin{align}
    & \underset{
       \mC,\mE,\mDelta,\mZ_C,\mZ_E}{\argmin} & & 
  \sum\limits_{k=1}^{K}\left[w_1  \norm{\text{vec}(\mX_k) - (\mB_k \odot \mA) \mC(k,:)^T }_2^2)\right]\nonumber \\
   & & & \ + w_2 \norm{\tY - \KOp{\mE, \mF,\mG}}_F^2 + g_C(\mZ_C)  + g_E(\mZ_E) \nonumber  \\
    & \ \ \ \ \ \text{s.t. }& & \mH_{\mC} \vecn(\mC) = \mH_{\mC}^{\Delta} \vecn(\mDelta),\nonumber\\
   & & &\mH_{\mE} \vecn(\mE) = \mH_{\mE}^{\Delta} \vecn(\mDelta),\nonumber\\
  & & & \mC = \mZ_C, \nonumber\\
  & & & \mE = \mZ_E, \nonumber
\end{align}}
by introducing split variables $\mZ_{C}\in\mathbb{R}^{K\times R_1}$ and $\mZ_{E}\in\mathbb{R}^{I_2\times R_2}$, and vectorizing the PARAFAC2 part of the objective function using the identity
$
  \text{vec}\left( \mA \mX \mB\right)=\left(\mB^T \odot \mA\right)\diag{X},  
$
for all $\mA \in \R^{m\times n}$, $\mB\in\R^{p\times q}$ and diagonal matrices $\mX\in \R^{n\times p}$.
The problem's augmented Lagrangian, incorporating dual variables $\mu_{Z_C},\mu_{Z_E}$ for constraints, dual variables $\mu_{\Delta_C},\mu_{\Delta_E}$ for linear coupling, and parameters $\rho_C$ and $\rho_E$ acting as a step-size, reads as follows:
{\small\begin{align}
L&\left(\mC,\mE,\mZ_C,\mZ_E,\mu_{Z_C},\mu_{Z_E},\mDelta,\mu_{\Delta_C},\mu_{\Delta_E} \right)=\nonumber\\
&w_1\sum\limits_{k=1}^{K}\left[ \norm{\text{vec}(\mX_k) - (\mB_k \odot \mA) \mC(k,:)^T }_2^2\right]+ g_C(\mZ_C) \nonumber\\
&+\frac{\rho_C}{2}\norm{\mC - \mZ_C + \mu_{Z_C}}^2_F\nonumber\\
&+  w_2 \norm{\tY - \KOp{\mE ,\mF,\mG}}_F^2  + g_E(\mZ_E) + \frac{\rho_E}{2}\norm{\mE - \mZ_E + \mu_{Z_E}}^2_F\nonumber\\
&+  \frac{\rho_E}{2} \norm{\mH_{\mE} \vecn(\mE) - \mH_{\mE}^{\Delta} \vecn(\mDelta) + \vecn(\mu_{\Delta_E})}_2^2 \nonumber\\ 
&+ \frac{\rho_C}{2} \norm{\mH_{\mC} \vecn(\mC) - \mH_{\mC}^{\Delta} \vecn(\mDelta) + \vecn(\mu_{\Delta_C})}_2^2. \label{eq:AugLagrC}
\end{align}}
For better readability, we restrict ourselves to exact couplings first, \textit{i.e.} couplings of the form:
\begin{equation*}
\mC = \mDelta,\ \ \ \mE = \mDelta 
\end{equation*}
Unlike the previously presented updates for the coupled mode $\mA$, the updates for coupled mode $\mC$ have to be executed row-wise for each $\mC(k,:)$, due to the structure of PARAFAC2.
In order to minimize the augmented Lagrangian \eqref{eq:AugLagrC} w.r.t. $\mC(k,:)$, the following linear system has to be solved,
{\small \begin{equation}\label{eq:C_update}
\begin{aligned}
&\mC(k,:)^{(n+1)} \left[\omega_1 (\mA^T\mA \ast \mB_k^T\mB_k)+\frac{\rho_{c_k}}{2}(\mI_{R_1}+\mI_{R_1})\right]
=\\
&\omega_1 \diag{\mA^T\mX_k\mB_k}^T\\
& \ \ \ \ \ + \frac{\rho_{c_k}}{2} \left[\mZ_{C}^{(n)}(k,:)-\vmu_{Z_C}^{(n)}(k,:)+\mDelta^{(n)}(k,:)-\vmu^{(n)}_{\Delta_C}(k,:) \right].
\end{aligned}
\end{equation}}
In the derivation of this, following transformations have been used \cite{KoBa09}:
\begin{align}
(\mB_k \odot \mA)^T(\mB_k \odot \mA)=\mA^T\mA \ast \mB_k^T\mB_k, \label{eq:KOptrick}\\
(\mB_k \odot \mA)^T\text{vec}(\mX_k) = \diag{\mA^T\mX_k\mB_k}.
\end{align}
In this way, we can avoid explicit computations of costly Khatri-Rao products and make the algorithm more efficient.
Following the argument in \cite{HuSiLi16,ScCoAc21}, we choose an individual step-size parameter $\rho_{c_k}$ for each $k=1,...,K$ and set $\rho_{c_k}=\text{trace}(\mA^T\mA \ast \mB_k^T\mB_k)/R_1$.
Minimizing the augmented Lagrangian \eqref{eq:AugLagrC} w.r.t. factor  matrix $\mE$ results in solving the linear system 
\begin{equation}\label{eq:E_update}
\begin{aligned} &\mE^{(n+1)} \left[w_2 \left(\mG^T \mG \ast \mF^T \mF\right) +  \frac{\rho_E}{2} \left(\mI_{R_2} + \mI_{R_2} \right) \right] =\\
	            		  &\left[ w_2 \tY_{\left[1\right]} \left(\mG \odot \mF \right) + \frac{\rho_E}{2}\left( \mZ_{E}^{(n)}-\vmu_{Z_{E}}^{(n)} + \mDelta^{(n)} - \vmu_{\mDelta_E}^{(n)}\right)\right],
\end{aligned}
\end{equation}
where we have again applied the relation \eqref{eq:KOptrick}. Here, we set $\rho_E=\text{trace}(\mG^T \mG \ast \mF^T \mF)/R_2$.
The complete ADMM algorithm for this subproblem in the case of exact coupling is given in Algorithm \ref{alg:admm_Cmode}. 
The stopping conditions are derived directly from \eqref{eq:ADMMstopcrit} and are provided in the supplementary material, together with more details about the algorithm.
Note that the update of $\mDelta$ is also computed row-wise due to different step-sizes $\rho_{c_k}$ for each $k$:
\begin{equation}\label{eq:Deltaupdate}
\begin{aligned}
    \mDelta^{(n+1)}(k,:) = \frac{1}{\rho_E+\rho_{c_k}}&\left[ \rho_E (\mE^{(n+1)}(k,:)+\vmu_{\Delta_{E}}^{(n)}(k,:))\right.\\
    + &\ \left. \rho_{c_k}(\mC^{(n+1)}(k,:)+\vmu_{\Delta_{C}}^{(n)}(k,:)) \right]
    \end{aligned}
\end{equation}
When updating $\mZ_C$, we employ the maximum over all $\rho_{c_k}$ in the proximal operator of the regularization function as it results in the minimal step size.
\begin{algorithm}[h!]
        \begin{algorithmic}[1]
            \WHILE{convergence criterion is not met}
\FOR{$\small{k=1,...,K}$}
	            \STATE{$\small{\mC^{(n+1)} (k,:)\longleftarrow \text{solve linear system } \eqref{eq:C_update}
	            		  }$}
\ENDFOR{}	
	            \STATE{$\small{\mE^{(n+1)} \longleftarrow \text{solve linear system } \eqref{eq:E_update}}$}
\FOR{$\small{k=1,...,K}$} 		 
        \STATE{$\small{\mDelta^{(n+1)}(k,:)\longleftarrow \text{given by } \eqref{eq:Deltaupdate}} $}	            		
\ENDFOR{}	
                \STATE{$\small{\begin{aligned} \mZ_{C}^{(n+1)}=
		            \prox_{\frac{1}{\max{\rho_{c_k}}}g_{C}}\left(\mC^{(n+1)}+\vmu_{Z_{C}}^{(n)}  \right)\end{aligned}} $}
		         \STATE{$\small{\begin{aligned} \mZ_{E}^{(n+1)}=
		            \prox_{\frac{1}{\rho_E}g_{E}}\left(\mE^{(n+1)}+\vmu_{Z_{E}}^{(n)}  \right)\end{aligned}} $}
             \STATE{$\small{ \vmu_{Z_{C}}^{(n+1)} = \vmu_{Z_{C}}^{(n)} +
	            \mC^{(n+1)} - \mZ_{C}^{(n+1)} }$}
	            \STATE{$\small{ \vmu_{Z_{E}}^{(n+1)} = \vmu_{Z_{E}}^{(n)} +
	            \mE^{(n+1)} - \mZ_{E}^{(n+1)} }$}
	            \STATE{$\small{ \vmu_{\mDelta_C}^{(n+1)} = \vmu_{\mDelta_C}^{(n)} + \mC^{(n+1)} - \mDelta^{(n+1)} }$}
	            \STATE{$\small{ \vmu_{\mDelta_E}^{(n+1)} = \vmu_{\mDelta_E}^{(n)} + \mE^{(n+1)} - \mDelta^{(n+1)} }$}

            \STATE{$\small{n = n+1}$}
            \ENDWHILE{}
        \end{algorithmic}
        \caption{ADMM for subproblem w.r.t. $\mC$ and $\mE$}
\label{alg:admm_Cmode}
\end{algorithm}
For the other instances of linear couplings as defined in \cite{ScCoAc21}, adjustments need to be made to the updates of $\mC, \mE,\mDelta,\vmu_{\mDelta_C}$ and $\vmu_{\mDelta_E}$ in Algorithm \ref{alg:admm_Cmode}. We provide the modified updates for $\mC$ and $\mDelta$ for each instance below. The corresponding updates for $\mE,\vmu_{\mDelta_C}$ and $\vmu_{\mDelta_E}$ are provided in the supplementary material. 
Coupling types $2a$ and $2b$ describe transformations in mode dimension which can, for instance, be used to account for different sampling grids between coupled datasets. Coupling types $3a$
 and $3b$, on the other hand, describe transformations in component dimension that can be used to couple decompositions that have both shared and unshared components. We refer to the supplementary material of \cite{ScCoAc21} for a discussion on rank and size restrictions on the particular transformation matrices in each instance.\\
\textbf{Case $2a$}
Here, the linear coupling takes the following form,
\begin{equation}
\tilde{\mH}_{\mC}\mC = \mDelta, \ \ \ \tilde{\mH}_{\mE}\mE = \mDelta.
\end{equation}
For this type of transformation, the augmented Lagrangian \eqref{eq:AugLagrC} is not separable in the rows of $\mC$. Therefore, minimizing \eqref{eq:AugLagrC} w.r.t. factor matrix $\mC$ results in solving a large linear system which is vectorized in $\mC$,
{\small\begin{equation*}
\begin{aligned}
&\left[\frac{\rho_C}{2}\left(\mI_{KR_1}+\left(\tilde{\mH}_{\mC} \otimes \mI_{R_1} \right)^T \left(\tilde{\mH}_{\mC} \otimes \mI_{R_1} \right)\right)\right.\\
& \left.+ \omega_1 \mM^T\mM \right]\text{vec}(\mC^T)=\\
& \frac{\rho_C}{2}\left[\text{vec}(\mZ_C^T - \vmu_{Z_C}^T)+ \left(\tilde{\mH}_{\mC} \otimes \mI_{R_1} \right)^T\text{vec}(\mDelta^T - \vmu_{\Delta_C}^T) \right]\\
& + \omega_1 \mM^T\text{vec}(\tX)
\end{aligned}
\end{equation*}}
where $\mM$ is a block-diagonal matrix with the matrices $(\mB_k \odot \mA)$, $k=1,...,K$, on its diagonal.
Here, we set $\rho_C=\text{trace}(\mM^T\mM)/(R_1K)$.
The update of $\mDelta$ is given by:
{\small\begin{equation*}
\begin{aligned}
\mDelta^{(n+1)} = \frac{1}{\rho_C+\rho_E}\Bigl[  & \rho_C \left(\tilde{\mH}_{\mC}\mC^{(n+1)}
+\vmu_{\mDelta_C}^{(n)} \right)\\
& \left.+\rho_E \left(\tilde{\mH}_{\mE}\mE^{(n+1)}+\vmu_{\mDelta_E}^{(n)} \right)\right]
\end{aligned}
\end{equation*}}\\
\textbf{Case $2b$}
In this case, the linear coupling is defined via
\begin{equation*}
\mC = \tilde{\mH}_{\mC}^{\mDelta}\mDelta, \ \ \ \mE = \tilde{\mH}_{\mE}^{\mDelta} \mDelta.
\end{equation*}
The update for the $k$-th row of $\mC$ is then given by the solution of the following system,
{\small\begin{equation*}\begin{aligned}
&\mC^{(n+1)}(k,:) \left[\omega_1 (\mA^T\mA \ast \mB_k^T\mB_k)+\frac{\rho_{c_k}}{2}(\mI_{R_1}+\mI_{R_1})\right]^T = \\
&\omega_1 \diag{\mA^T\mX_k\mB_k}^T\\
& \ \ \ \ \ + \frac{\rho_{c_k}}{2} \left[\mZ_{C}^{(n)}(k,:)-\vmu_{Z_{C}}^{(n)}(k,:) +\tilde{\mH}_{\mC}^{\mDelta}(k,:) \mDelta^{(n)}-\vmu_{\Delta_{C}}^{(n)}(k,:) \right],
\end{aligned}
\end{equation*}}
and the update for $\mDelta$ is given by
{\small\begin{equation*}\begin{aligned}
&\left(\rho_E \tilde{\mH}_{\mE}^{\mDelta^T}\tilde{\mH}_{\mE}^{\mDelta} + \tilde{\mH}_{\mC}^{\mDelta^T} \mP \tilde{\mH}_{\mC}^{\mDelta} \right) \mDelta = \\
&\rho_E \tilde{\mH}_{\mE}^{\mDelta^T}\left(\mE^{(n+1)}+\vmu_{\Delta_E}^{(n)}\right) + \tilde{\mH}_{\mC}^{\mDelta^T}\mP \left(\mC^{(n+1)} + \vmu_{\Delta_{C}}^{(n)} \right),
\end{aligned}
\end{equation*}}
where
\begin{equation*}
\mP = \left[\begin{array}{ccc} 
\rho_{c_1} & & \mathbf{0}\\
& \ddots & \\
\mathbf{0} & & \rho_{c_K}
\end{array} \right]
\end{equation*}
Here, $\rho_{c_k}$ is defined as in case $1$.\\
\textbf{Case $3a$}
Here, linear couplings of the form
\begin{equation*}
\mC \hat{\mH}_{\mC} = \mDelta, \ \ \ \mE \hat{\mH}_{\mE} = \mDelta
\end{equation*}
are considered.
In order to update the $k$-th row of $\mC$ the following system needs to be solved:
{\small\begin{equation*}\begin{aligned}
&\mC^{(n+1)}(k,:) \left[\omega_1 (\mA^T\mA \ast \mB_k^T\mB_k)+\frac{\rho_{c_k}}{2}(\mI_{R_1}+\hat{\mH}_{\mC}\hat{\mH}_{\mC}^T)\right]
= \\
&\ \ \ \ \omega_1 \diag{\mA^T\mX_k\mB_k}^T\\
&+ \frac{\rho_{c_k}}{2} \left[\mZ_{C}^{(n)}(k,:)-\vmu_{Z_{C}}^{(n)}(k,:)+\left(\mDelta^{(n)}(k,:)-\vmu_{\Delta_{C}}^{(n)}(k,:)\right)\hat{\mH}_{\mC}^T \right]
\end{aligned}
\end{equation*}}
Also the update of $\mDelta$ is given row-wise via
{\small\begin{equation*}\begin{aligned}
\mDelta^{(n+1)}(k,:) = \frac{1}{\rho_E+\rho_{c_k}}&\Bigl[\rho_E (\mE^{(n+1)}(k,:)\hat{\mH}_{\mE}+\vmu_{\Delta_E}^{(n)}(k,:))\\
& \left.+\rho_{c_k}(\mC^{(n+1)}(k,:)\hat{\mH}_{\mC}+\vmu_{\Delta_{C}}^{(n)}(k,:)) \right]
\end{aligned}
\end{equation*}}\\
\textbf{Case $3b$}
Given linear couplings of the form
\begin{equation*}
\mC = \mDelta \hat{\mH}_{\mC}^{\mDelta}, \ \ \ \mE = \mDelta \hat{\mH}_{\mE}^{\mDelta},
\end{equation*}
the following system has to be solved to update the $k$-th row of $\mC$:
{\small\begin{equation*}\begin{aligned}
&\mC^{(n+1)}(k,:) \left[\omega_1 (\mA^T\mA \ast \mB_k^T\mB_k)+\frac{\rho_{c_k}}{2}(\mI_{R_1}+\mI_{R_1})\right] = \\
&\omega_1 \diag{\mA^T\mX_k\mB_k}^T\\
&\ \ \ + \frac{\rho_{c_k}}{2} \left[\mZ_{C}^{(n)}(k,:)-\vmu_{Z_{C}}^{(n)}(k,:) +\mDelta^{(n)}(k,:)\hat{\mH}_{\mC}^{\mDelta}-\vmu_{\mDelta_{C}}^{(n)}(k,:) \right]
\end{aligned}
\end{equation*}}
The update of $\mDelta$ is given via
{\small\begin{equation*}
\begin{aligned}
 &\mDelta^{(n+1)}(k,:) \left( \rho_E \hat{\mH}_{\mE}^{{\mDelta}}\hat{\mH}_{\mE}^{\mDelta^T} + \rho_{c_k} \hat{\mH}_{\mC}^{{\mDelta}}\hat{\mH}_{\mC}^{\mDelta^T}\right) =  \\
& \ \ \ \ \rho_E  \left(\mE^{(n+1)}(k,:)+\vmu_{\mDelta_E}^{(n)}(k,:) \right)\hat{\mH}_{\mE}^{\mDelta^T}\\
&+\rho_{c_k} \left(\mC^{(n+1)}(k,:)+\vmu_{\mDelta_{C}}^{(n)}(k,:) \right)\hat{\mH}_{\mC}^{\mDelta^T}
 \end{aligned}
\end{equation*}}

\subsection{Computational complexity and efficient implementations}
The single most expensive step of the AO-ADMM algorithm is the computation of the matricized tensor times Khatri-Rao product ${\tY}_{[1]}(\mG \odot \mF)$ in \eqref{eq:E_update}, when a CP decomposition is involved in the model, as in \eqref{eq:costfuncC}. Computing ${\tY}_{[1]}(\mG \odot \mF)$ has a worst-case complexity of $\mathcal{O}(I_2LPR_2)$, but it can be computed more efficiently via the \texttt{mttkrp} function from the Tensor Toolbox \cite{Tensortoolbox}. Furthermore, the product is precomputed outside the ADMM loop of Algorithm \ref{alg:admm_Cmode} and thus has to be computed only once per outer AO iteration.

Another expensive step in the algorithm is the solution of $K$ orthogonal Procrustes problems involving the computation of $K$ singular value decompositions of matrices of size $J_k\times R_1$, in the update of factor matrices $\{\mB_k\}_{k=1}^K$, with a cost of roughly $\mathcal{O}(\sum_{k=1}^K J_k R_1^2)$, see the supplementary material for the detailed update. This step is especially expensive as it has to be computed in every inner ADMM iteration.

The complexity of the update of factor matrix $\mC$ in model $\eqref{eq:costfuncC}$ depends on the type of linear coupling.
For all types of couplings, except coupling case $2a$, the update of the $k$-th row of the factor matrix $\mC$ is given by the solution of a linear system where the matrix inverse is only of size $R_1 \times R_1$. However, these $K$ matrix inverses are never explicitly computed. Instead, Cholesky decompositions of those matrices are precomputed outside the ADMM loop, which costs $\mathcal{O}(KR_1^3)$. Thus,
solving the linear systems at each ADMM iteration reduces to one forward- and one backward-substitution with complexity $\mathcal{O}(KR_1^2)$. Note, however, that both the $K$ Cholesky decompositions and the $K$ row-updates of $\mC(k,:)$ can be computed in parallel. In coupling case $2a$, one large, but sparse, linear system of size $KR_1 \times KR_1$ has to be solved in every ADMM iteration. Also in this case, we precompute a Cholesky decomposition (with cost $\mathcal{O}(K^3R_1^3)$ for a full matrix). Then the forward- and backward-substitution per ADMM iteration has a worst-case complexity $\mathcal{O}(K^2R_1^2)$. However, in practice the complexity is often lower, since sparse computations are used.

Also for the update of $\mDelta$, a Cholesky decomposition of the matrices $\rho_E \tilde{\mH}_{\mE}^{\mDelta^T}\tilde{\mH}_{\mE}^{\mDelta} + \tilde{\mH}_{\mC}^{\mDelta^T} \mP \tilde{\mH}_{\mC}^{\mDelta}$ and $  \rho_E \hat{\mH}_{\mE}^{{\mDelta}}\hat{\mH}_{\mE}^{\mDelta^T} + \rho_{c_k} \hat{\mH}_{\mC}^{{\mDelta}}\hat{\mH}_{\mC}^{\mDelta^T}$, $k=1,..,K$ in cases $2b$ and $3b$ respectively, can be precomputed before the ADMM iterations. In case $2b$, the Cholesky decomposition has a complexity of $\textit{O}({m_1}^3)$ and each forward and backward-substitution $\textit{O}({m_1}^2 R_1)$. In case $3b$, it is $\textit{O}(Km_2^3)$ and $\textit{O}(Km_2^2)$, respectively. In cases $1$, $2a$ and $3a$, the update of $\mDelta$ is computed as an average with complexity $\textit{O}(KR_1)$, $\textit{O}(m_1R_1)$ and $\textit{O}(K\max{\{R_1,R_2\}}m_2)$, respectively. Note, again, that row-wise updates can be done in parallel.

The complexities of the updates of modes $\mB_k$ and $\mA$ of model \eqref{eq:costfuncC} are provided in the supplementary material. 
In all cases, there are of course additional costs that result from assembling matrices and vectors involved in the updates. However, these depend very much on the setting of couplings and constraints at hand. In this connection, note that the products $\mA^T \mA, \mB_k^T \mB_k, \mC^T\mC,...$ are precomputed and stored throughout the whole AO-ADMM algorithm. They only need to be updated for each mode after the corresponding outer AO iteration.

Finally, we use AO with so-called \textit{warm starts} as proposed in \cite{HuSiLi16}. That means, we initialize each ADMM algorithm with the values from the previous AO iteration.

The complexity of the update of split variables $\mZ$ depends on the proximal operator. For many commonly used constraints, closed form solutions or efficient algorithms exist. In the case of non-negativity, which is the only possible constraint in TASTE and C$^3$APTION, the proximal operator is simply given by elementwise thresholding at zero, $\max \left\lbrace 0,z_{i,j}\right\rbrace$, so the cost is only linear in the size of the matrix.

TASTE and C$^3$APTION, however, do not use ADMM and proximal operators. Instead they rely on solving the NNLS problems directly via solvers like BPP or active set. Therefore, the efficiency of those algorithms mainly depends on the efficiency of the NNLS solver. When a CP decomposition is involved in the model, as in C$^3$APTION \eqref{eq:costfuncCAPTION}, the same matricized tensor times Khatri-Rao product as in AO-ADMM has to be computed once per outer AO iteration. Also the orthogonal Procrustes problems involved in the update of the matrices $\mB_k$ are almost the same as in the AO-ADMM framework and have therefore the same complexity. The difference, however, is that those updates are only done once in one outer AO iteration of TASTE and C$^3$APTION, while in AO-ADMM, they are performed in every inner ADMM iteration of the mode $\mB$ update. Generally, AO-ADMM is more complex than TASTE and C$^3$APTION. It uses also more variables (dual and split variables) and therefore more memory. On the other hand, AO-ADMM is built to be flexible with respect to couplings and constraints/regularizations. The following experiments even show that the difference in per iteration complexity between the algorithms is rather small, while AO-ADMM generally needs less outer iterations to reach the same accuracy as TASTE and C$^3$APTION.

\section{Experiments}\label{sec:experiments}
Here, we present numerical experiments on synthetic data (Experiments $1-4$) and demonstrate the use of a PARAFAC2-based CMTF model in a novel metabolomics application. Experiments $1$ and $2$ show the comparisons with state-of-the-art methods TASTE and C$^3$APTION, respectively, in a setting with non-negativity constraints and exact coupling in mode $\mC$ and varying levels of difficulty given by noise levels, data sizes and number of components. Experiment $3$ demonstrates the fusion of dynamic data with evolving patterns together with static data and Experiment $4$ shows the flexibility of the framework in a setting with smoothness regularization and partial couplings.

\subsection{Synthetic Data}
\subsubsection{Experimental Set-Up}
For each experiment, we generate $20$ random coupled dataset pairs with tensors $\tX$ and $\tY$, where $\tX$ and $\tY$ are constructed from known ground-truth factor matrices following either a PARAFAC2-, a CP- or a matrix decomposition. Unless specified otherwise, factor matrices $\mB_k$, $k=1,...,K$, are generated as shifted versions of each other and thus fulfil the PARAFAC2 constraint. Noise is then added to each data tensor as 
\begin{equation*}
    {\tX}_{\textrm{noisy}} = {\tX}+\eta\tN \left( \norm{\tX}_{F}/ \norm{\tN}_{F}\right),
\end{equation*}
where $\small{\tN}$ is a noise tensor with entries drawn from the standard normal distribution and $\eta$ specifies the noise level. Furthermore, each data tensor is normalized to Frobenius norm $1$ and weighted equally, \textit{i.e.} $w_i$ in \eqref{eq:costfuncA},\eqref{eq:costfuncC} are set to $0.5$.
When fitting the models to each dataset pair, we use $10$ random initializations (unless specified otherwise), and report the results only for the run with the lowest cost function value, \textit{i.e.} \eqref{eq:func_value}.
Factor matrices are initialized with entries drawn from the standard normal or, in the case of non-negative factors, uniform distribution, after which the columns are normalized. All split and dual variables are initialized by drawing from the uniform distribution.
As stopping conditions, we set all tolerances of the inner ADMM loops to be $10^{-5}$, the maximum number of inner iterations to $5$, and the outer absolute and relative tolerances to be $10^{-7}$ and $10^{-8}$, respectively.
We evaluate the performance of different algorithms in terms of the achieved function value, model fit, PARAFAC2 residual and the factor match score (FMS). 
The function value is defined as the regularized sum of squared errors, \textit{e.g.} for the model \eqref{eq:costfuncA}, the function value is given by the term
{\small\begin{equation}\label{eq:func_value}\begin{aligned}
    &\sum\limits_{k=1}^{K}\left[w_1  \norm{\smash{\mX_k  -  \mA \mD_k \mB_k^T }}_F^2  +  g_D( \mD_k )  +  g_B(\mB_k)\right]\\
    &+  w_2 \norm{\smash{\mY  -  \mE \mF^T}}_F^2  +  g_A(\mA)  +  g_E(\mE)  +  g_F(\mF).
\end{aligned}\end{equation}}
The model fit measures the reconstruction error of a single decomposition and is computed as 
\begin{equation*}
    \text{Fit}=100 \times \left(1-\norm{\smash{\tZ -\tilde{\tZ}}}_F^2 / \norm{\tZ}_F^2 \right),
\end{equation*}
where tensor $\tilde{\tZ}$ denotes the reconstructed version of $\tZ$.
The PARAFAC2 coupling residual is a measure of how well the computed factor matrices adhere to the PARAFAC2 constraint \eqref{eq:PAR2constralt} and is defined as follows,
\begin{equation*}
\frac{1}{K} \sum\limits_{k=1}^{K}\left(\| \mB_k^{(n)} - \mP_k^{(n)}\mDelta_B^{(n)}\|_F \big/ \|\mB_k^{(n)} \|_F\right).   
\end{equation*}
For a perfect PARAFAC2 model, the PARAFAC2 coupling residual would be zero.
Finally, the FMS measures the accuracy of the recovered components, and for an $R$-component PARAFAC2 model it is defined as
\begin{equation*}
    \text{FMS}=\frac{1}{R}\sum\limits_{r=1}^{R}\frac{\langle \va_r,\hat{\va}_r \rangle}{\norm{\va}_2\norm{\hat{\va}_r}_2} 
    \frac{\langle \vb_r,\hat{\vb}_r \rangle}{\norm{\vb}_2\norm{\smash{\hat{\vb}_r}}_2}
    \frac{\langle \vc_r,\hat{\vc}_r \rangle}{\norm{\vc}_2\norm{\hat{\vc}_r}_2},
\end{equation*}
Here, $\va_r$ and $\hat{\va}_r$ correspond to the $r$th column of the ground-truth factor matrix $\mA$ and recovered matrix $\hat{\mA}$, respectively (equivalently also for factor matrix $\mC$). For mode $\mB$, the vector $\vb_r$ contains the concatenation of the $r$-th columns of all $\mB_k$ matrices. Before computing the FMS, we find the best matching permutation of the components. The FMS of a CP- or matrix-decomposition is defined correspondingly and the total FMS of a coupled model is given by the product of the FMSs of the individual decompositions. We also evaluate the FMS of an individual mode by only considering the relevant terms, \textit{e.g.} for mode $\mA$
\begin{equation*}
    \text{FMS}_{\text{A}}=\frac{1}{R}\sum\limits_{r=1}^{R}\langle \va_r,\hat{\va}_r \rangle / \left(\norm{\va}_2\norm{\hat{\va}_r}_2\right).
\end{equation*}
The experiments were carried out with the following computer configurations: CPU: Intel Core i9-14900KF @ 3.40Hz; Memory: 128.00 Gb; System: 64-bit Windows 11; Matlab R2022b. 

\subsubsection{Comparison with TASTE}
We compare our proposed AO-ADMM algorithm with the TASTE algorithm described in Section \ref{sec:TASTE} on simulated datasets in three different settings. The dataset pairs consist of a tensor $\tX$ and a matrix $\mY$. The ground-truth factor matrices have rank $R=4$ 
and are non-negative. Entries in factor matrices $\mA$ and $\mB_k$ are drawn from the uniform distribution on $\left[0,1\right]$ and the entries in $\mC$ from the uniform distribution on $\left[0.1,1.1\right]$ to avoid near-zero elements which make the recovery of $\mB_k$ more difficult. For a detailed discussion on near-zero elements in $\mC$, we refer to \cite{RoScCa22}, which shows how the recovery of the $r$-th component of $\mB_k$ is directly affected by the noise in the data scaled by $1/\mC(k,r)$. This means that very small values in $\mC(k,r)$ lead to poor estimation of the corresponding component in $\mB_k$. Furthermore, the columnwise signal-to-noise ratio (cwSNR) has been defined and shown to be a good predictor of the accuracy of $\mB_k$ estimates providing the practitioner with a measure of quality of the solution. We then use AO-ADMM to fit the following model with exact coupling between $\mC$ and $\mE$ and non-negativity constraints $\iota_{\R_{\geq 0}}$ on all modes:
\begin{equation*}\small{
\begin{aligned}
    & \underset{\underset{
       \mA,\mE,\mF,\mDelta}{\left\lbrace \mB_k\right\rbrace_{k},\mC}}{\argmin} & & 
  \sum\limits_{k=1}^{K}\left[\frac{1}{2} \norm{\smash{\mX_k - \mA \diag{\mC(k,:)} \mB_k^T}}_F^2 + \iota_{\R_{\geq 0}}(\mB_k)\right] \\
   & & & \ + \frac{1}{2} \norm{\smash{\mY - \mE \mF^T}}_F^2 + \iota_{\R_{\geq 0}}(\mA) + \iota_{\R_{\geq 0}}(\mE)\\
   & & & + \iota_{\R_{\geq 0}}(\mC) + \iota_{\R_{\geq 0}}(\mF)\\
   & \ \ \ \ \ \text{s.t. } & & \left\lbrace\mB_k\right\rbrace_{k\leq K} \in \mathcal{P},\\
   & & & \mC = \mDelta,\mE  = \mDelta
\end{aligned}}
\end{equation*}
Using TASTE, we fit the same model, except that variables $\mC,\mE$ and $\mDelta$ are replaced by just one variable and the PARAFAC2 constraint is handled as a regularization term, see \eqref{eq:costfuncTASTE}. In \cite{AfPePa20}, the regularization parameter $\mu$ is set to $1$ and so, here, we set $\mu=1/\norm{\smash{\tX}}_F^{\frac{2}{3}}$ which corresponds to rescaling the factor matrix $\mB_k$ in order to compensate for the normalization of the data tensor. The code for TASTE is taken from \footnote{\url{https://github.com/aafshar/TASTE}}.

\begin{figure}[t]
\centering
\includegraphics[width=\columnwidth]{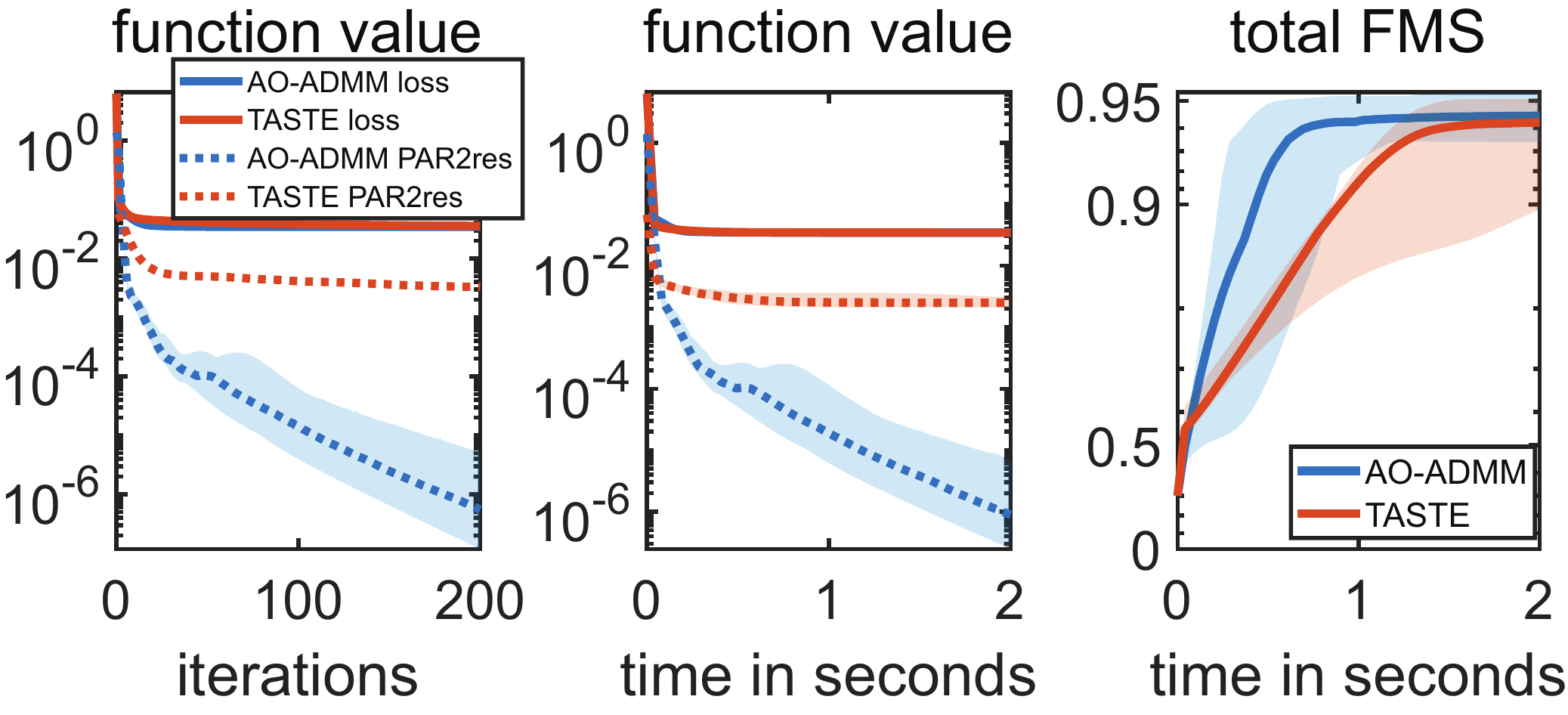}
\caption{Exp.~1a: Convergence comparison of AO-ADMM and TASTE.}
\label{fig:compTASTE1a}
\end{figure}

\paragraph{Experiment 1a: small sizes, few components, low noise}
Here, we have rather small datasets with a tensor $\tX$ of size $40 \times 60 \times 50$ and a matrix $\mY$ of size $50 \times 60$. The noise level is set to $0.2$ for both.
Figure~\ref{fig:compTASTE1a} shows the convergence behavior of both algorithms over iteration numbers and computation time. Note that no parallel computations have been used for any of the experiments. 
The (dotted) lines show the median performance over the $20$ datasets, while the semi-transparent area shows the minimum and maximum performance.
We see that both algorithms reach approximately the same accuracy in terms of FMS and AO-ADMM is slightly faster, even without the use of parallel computations. We also observe that the PARAFAC2 coupling residual is much smaller in AO-ADMM, which means that the AO-ADMM solution is much closer to a perfect PARAFAC2 structure compared to TASTE. Our experiments have also shown that other choices of the (constant) penalty parameter $\mu$ in TASTE lead to worse performance. This shows that the tuning of this hyper-parameter is important, but it is cumbersome in practice.

\begin{figure}[t]
\centering
\includegraphics[width=\columnwidth]{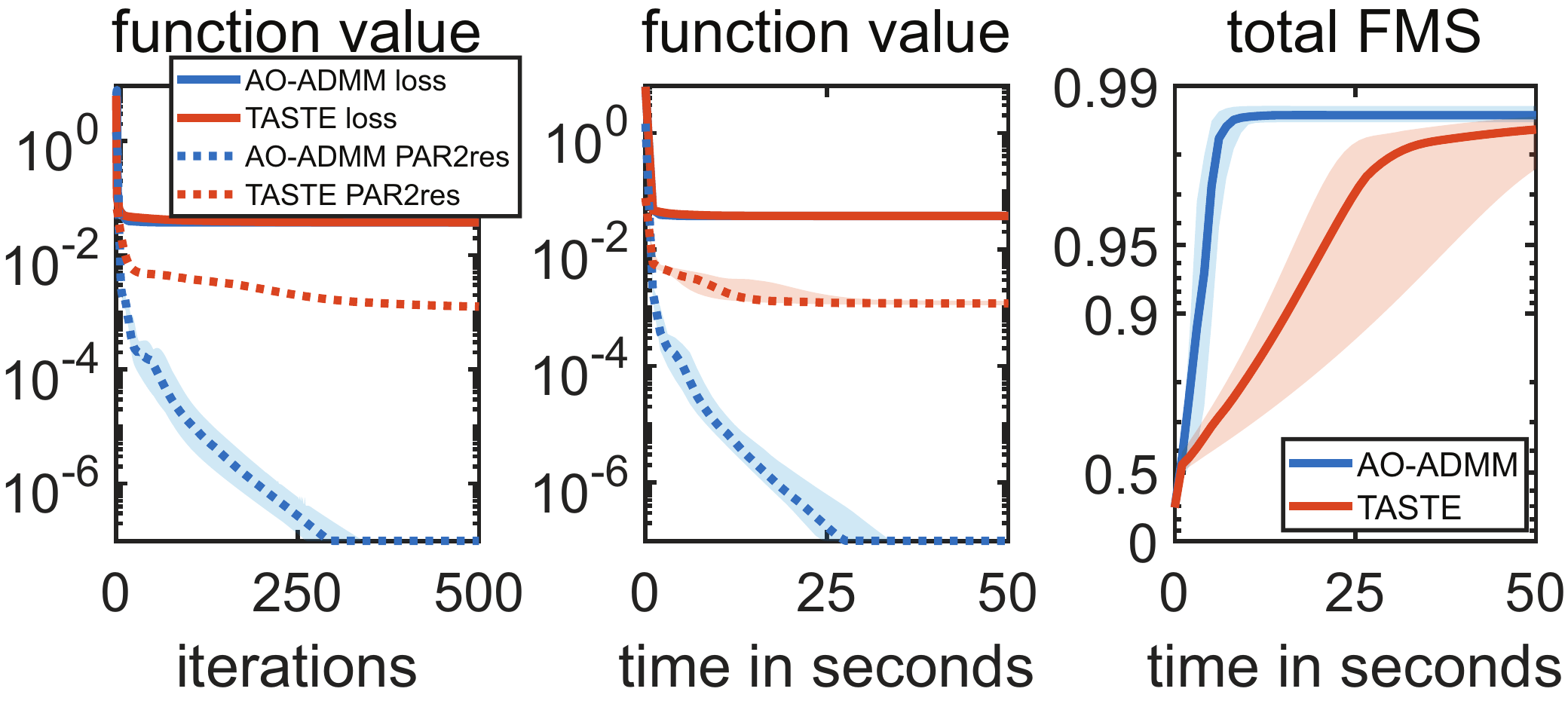}
\caption{Exp.~1b: Convergence comparison of AO-ADMM and TASTE.}
\label{fig:compTASTE1b}
\end{figure}

\begin{figure}[t]
\centering
\includegraphics[width=\columnwidth]{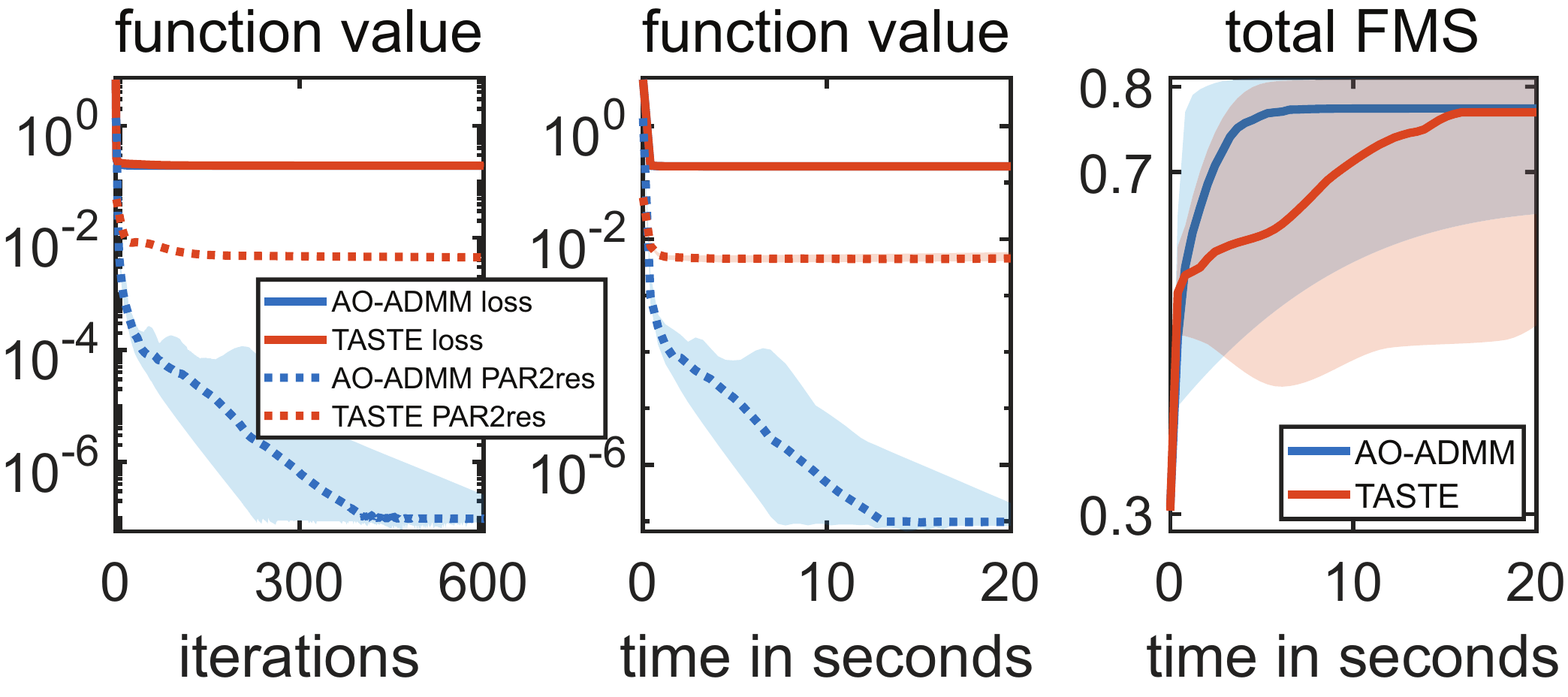}
\caption{Exp.~1c: Convergence comparison of AO-ADMM and TASTE.}
\label{fig:compTASTE1c}
\end{figure}

\paragraph{Experiment 1b: larger sizes}
In this setting, we have a tensor $\tX$ of size $200 \times 250 \times 200$ and a matrix $\mY$ of size $200 \times 300$. Otherwise, the setup is the same. Here, the difference in computation time becomes much more evident, as shown in Figure~\ref{fig:compTASTE1b}, with AO-ADMM reaching its final accuracy roughly $10$ times faster than TASTE. The overall accuracy in terms of FMS is higher here than in the previous experiment since we have now more data points (larger tensors) to estimate the same number of parameters in the decomposition.
\paragraph{Experiment 1c: more noise}
Here, we use again the small data sizes as in Experiment $1a$, but set the noise level for tensor $\tX$ to $0.8$. Figure~\ref{fig:compTASTE1c} shows that the accuracy achieved by both algorithms is lower than before, which is to be expected as this is a more difficult problem. We also observe that TASTE struggles more with the noise than AO-ADMM, indicated by a longer median computation time and a larger variation in computation time between datasets.

\subsubsection{Comparison with C$^3$APTION}
Here, we compare AO-ADMM with the three versions of the C$^3$APTION algorithm described in \ref{sec:CAPTION} on synthetically generated datasets consisting of a PARAFAC2-tensor $\tX$ and a CP-tensor $\tY$ in the same settings as before and an additional setting with a higher number of components. The factor matrices have again rank $R=4$ (except $R=10$ in the last setting) and are generated as described above. We denote the three versions of C$^3$APTION as C$^3$APTION-BPP, C$^3$APTION-ASET and C$^3$APTION-ALS, employing block principal pivoting, active set and alternating least-squares with zero-thresholding, respectively. The code for C$^3$APTION is taken from \footnote{\url{http://www.cs.ucr.edu/~egujr001/ucr/madlab/src/caption\_code.zip}}. We set again $\mu=1/\norm{\tX}_F^{\frac{2}{3}}$.

\begin{figure}[t]
\centering
\includegraphics[width=\columnwidth]{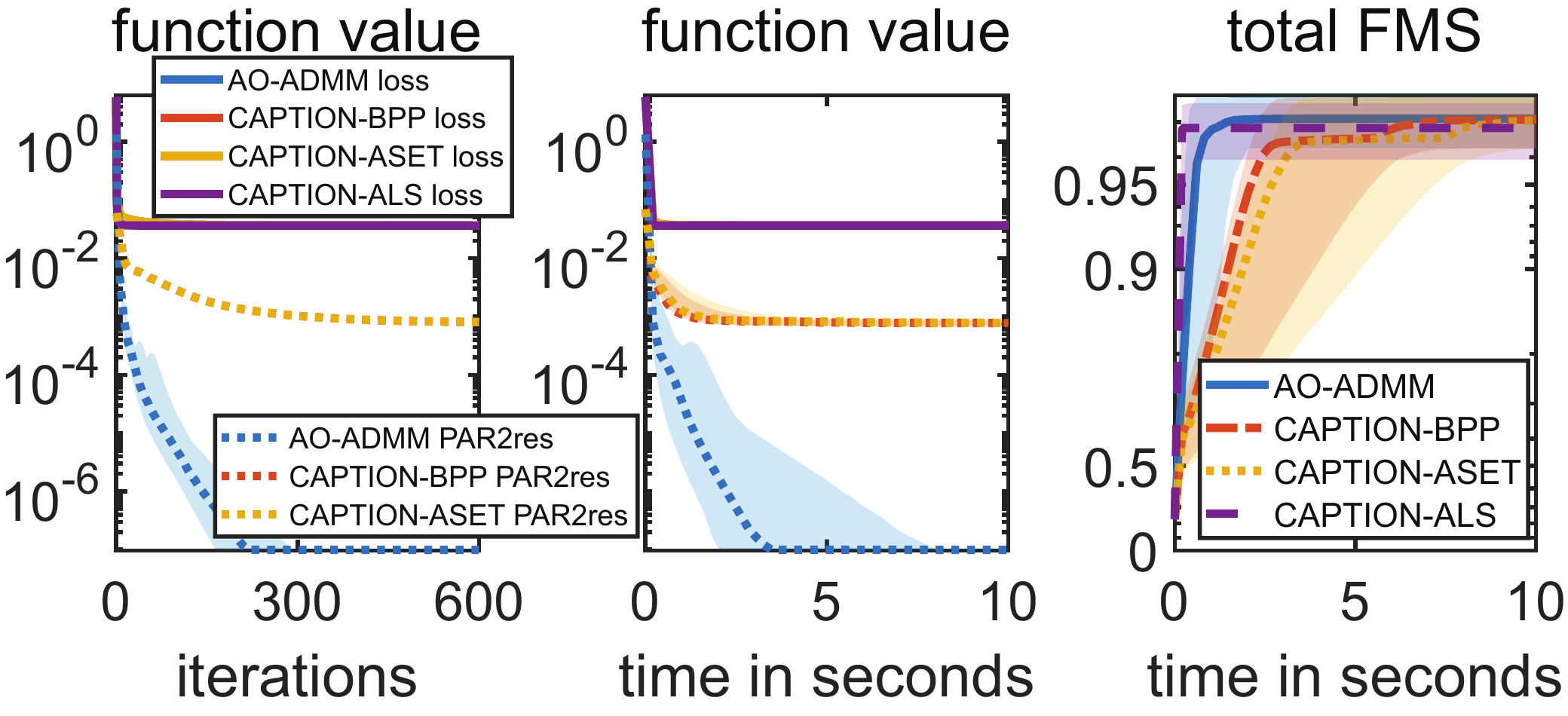}
\caption{Exp.~2a: Convergence comparison of AO-ADMM and different versions of C$^3$APTION.}
\label{fig:compCAPTION2a}
\end{figure}

\paragraph{Experiment 2a: small sizes, few components, low noise}
Here, tensor $\tX$ is of size $40 \times 60 \times 50$ and tensor $\tY$ of size $50 \times 60 \times 50$ and the noise level is $0.2$.
The comparison of the performance of all four algorithms is shown in Figure~\ref{fig:compCAPTION2a}.
We observe that C$^3$APTION-BPP and C$^3$APTION-ASET reach the same accuracy as AO-ADMM in terms of FMS but are slower. C$^3$APTION-ALS, on the other hand, is the fastest of all the algorithms, even slightly faster than AO-ADMM, but does not reach the same accuracy.
Note that C$^3$APTION-ALS makes use of ALS with thresholding at zero, which is fast, but uses a very rough approximation to the solution of NNLS subproblems and has no convergence guarantees \cite{KiHePa14}. This is a possible reason for the less accurate results of C$^3$APTION-ALS in this experiment compared to all other algorithms. C$^3$APTION-ALS also fails in one run due to zeroing-out a whole factor matrix.
Figure~\ref{fig:compCAPTION2a} also shows that AO-ADMM achieves a much smaller PARAFAC2 coupling residual compared to C$^3$APTION-BPP and C$^3$APTION-ASET, for which the convergence curves of the PARAFAC2 coupling residual coincide.

\begin{figure}[t]
\centering
\includegraphics[width=\columnwidth]{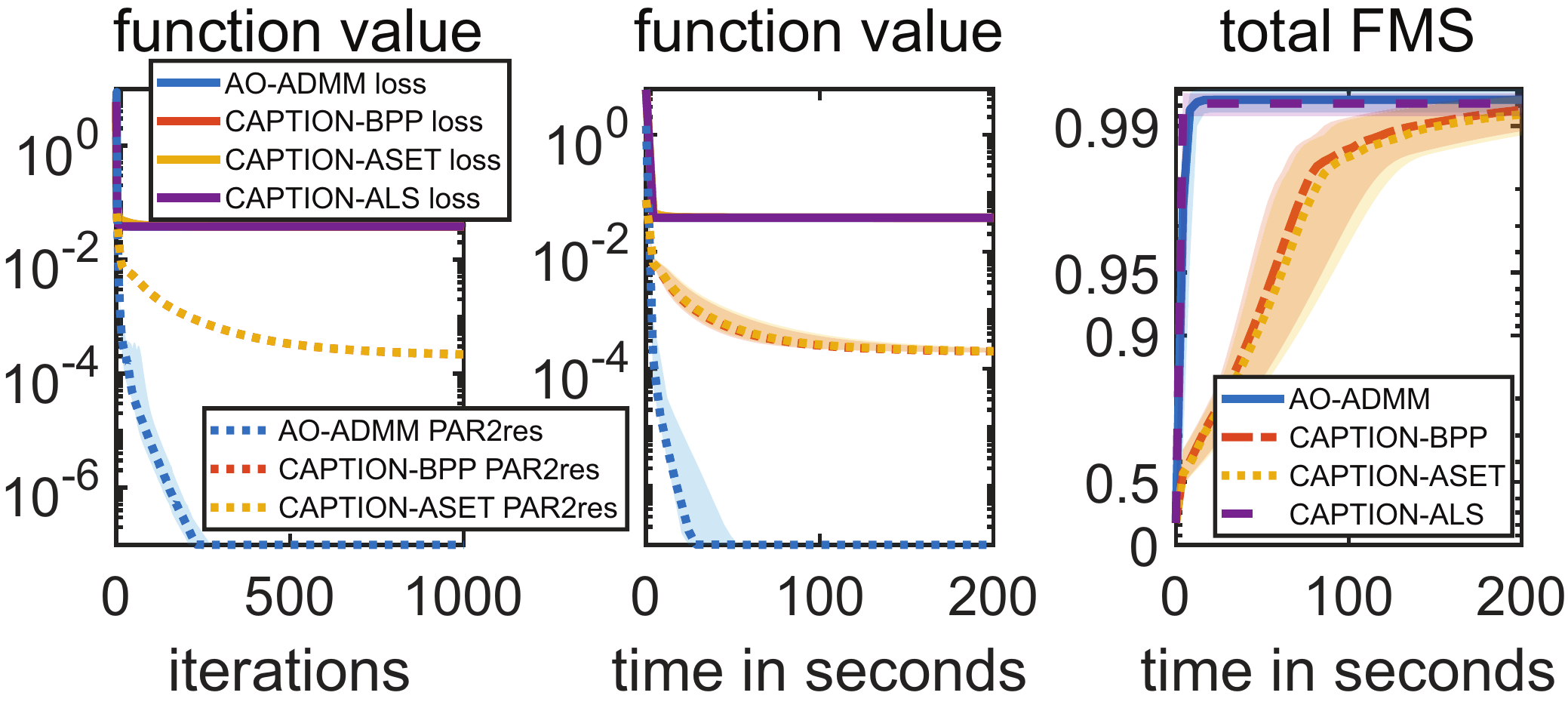}
\caption{Exp.~2b: Convergence comparison of AO-ADMM and C$^3$APTION.}
\label{fig:compCAPTION2b}
\end{figure}

\begin{figure}[t]
\centering
\includegraphics[width=\columnwidth]{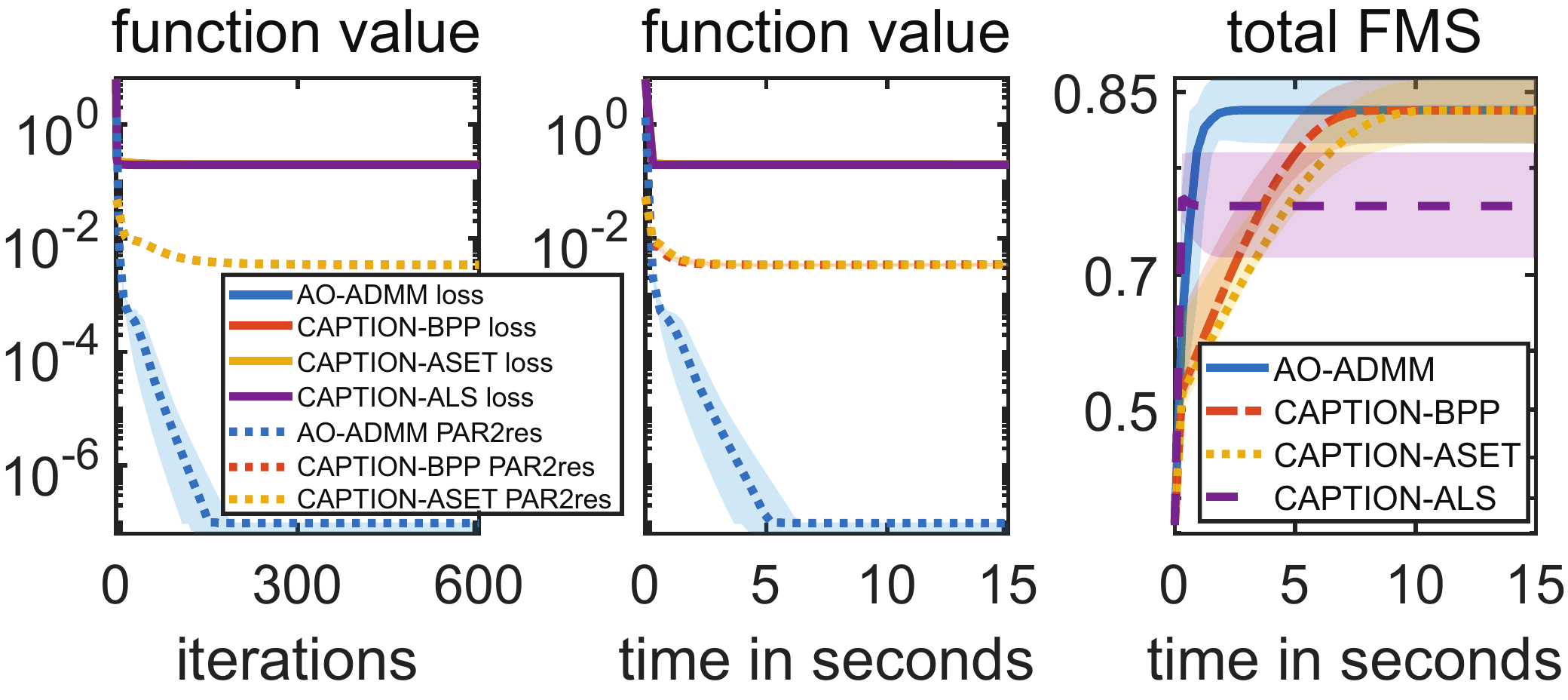}
\caption{Exp.~2c: Convergence comparison of AO-ADMM and C$^3$APTION.}
\label{fig:compCAPTION2c}
\vspace{-0.4cm}
\end{figure}

\begin{figure}[t]
\centering
\includegraphics[width=\columnwidth]{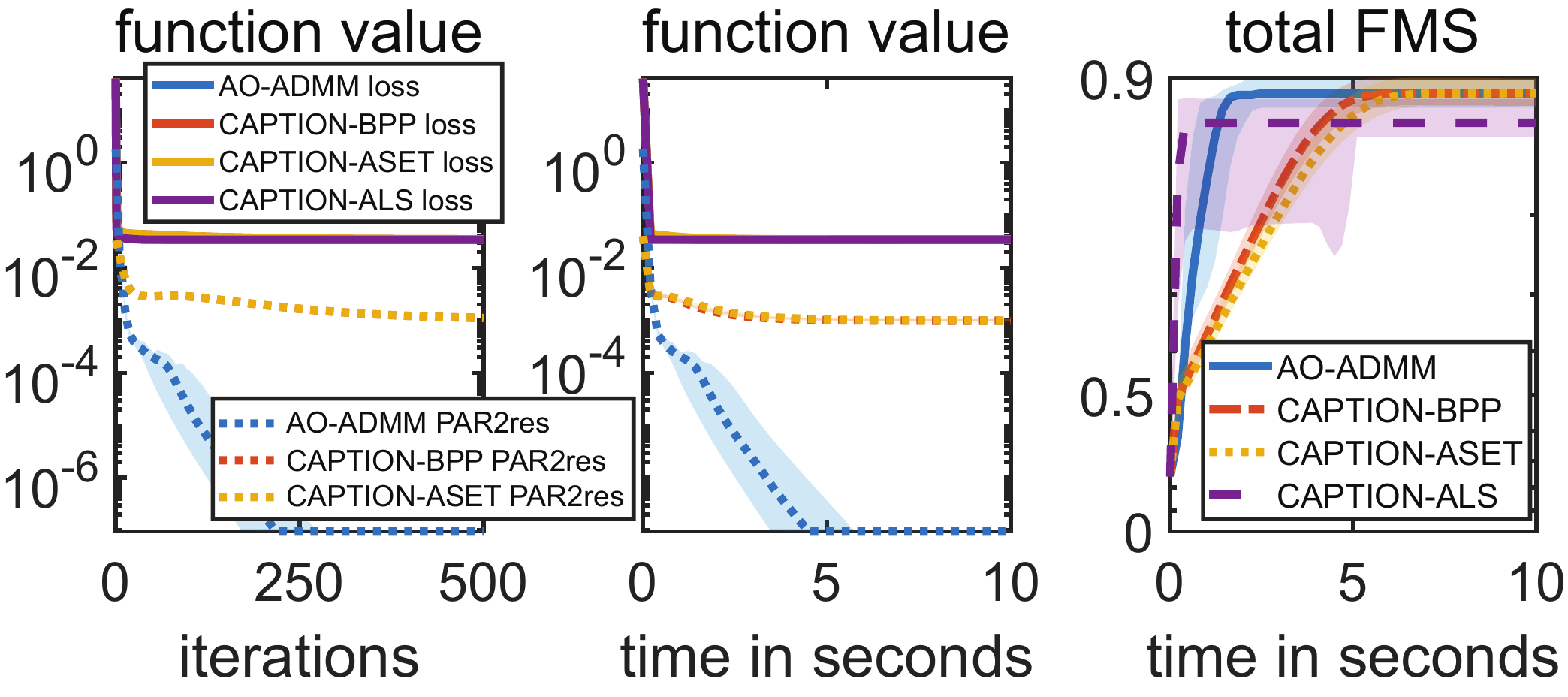}
\caption{Exp.~2d: Convergence comparison of AO-ADMM and C$^3$APTION.}
\label{fig:compCAPTION2d}
\vspace*{-4mm}
\end{figure}

\paragraph{Experiment 2b: larger sizes}
In this part, we construct a tensor $\tX$ of size $200 \times 250 \times 200$ and a tensor $\tY$ of size $200 \times 300 \times 200$, and other settings are the same as Experiment $2a$. Figure~\ref{fig:compCAPTION2b} shows that although all algorithms achieve a similar final accuracy, CAPTION-BPP and CAPTION-ASET are nearly $20$ times slower than the other two, which is more obvious than in Experiment $2a$. Note, that the convergence curves of the PARAFAC2 coupling residual of C$^3$APTION-BPP and C$^3$APTION-ASET coincide again. 
\paragraph{Experiment 2c: more noise}
Here, we keep the small tensors as in Experiment $2a$, but set the noise level for tensor $\tX$ to $0.8$. Figure~\ref{fig:compCAPTION2c} shows that while the relative performance of AO-ADMM compared to C$^3$APTION-BPP and C$^3$APTION-ASET is similar to the setting with low noise, the accuracy of C$^3$APTION-ALS is significantly lower with noisier data.
\paragraph{Experiment 2d: more components}
In this last experiment, we set the number of components to be $10$, while again using small tensors and low noise. The results are shown in Figure~\ref{fig:compCAPTION2d}. The relative performance of all methods stays roughly the same as in Experiment $2a$ with less components. The FMS achieved by all methods is lower than with less components. This is due to the fact that, here, we need to estimate a higher number of parameters with the same amount of data points as in Experiment $2a$.

\subsubsection{Experiment 3: Fusion of dynamic and static data}
In this example, we jointly analyze synthetically generated dynamic and static datasets using a PARAFAC2-based CMTF model with coupling in mode $\mA$, where PARAFAC2 is used to capture evolving patterns from the dynamic data. We demonstrate that the model can reveal evolving patterns accurately, even when those patterns do not exactly follow the PARAFAC2 constraint, which is a rather realistic assumption for real-world datasets, and that coupling with a less noisy dataset can improve the performance of downstream tasks such as clustering. 

\begin{figure}[t]
\centering
\includegraphics[width=\columnwidth]{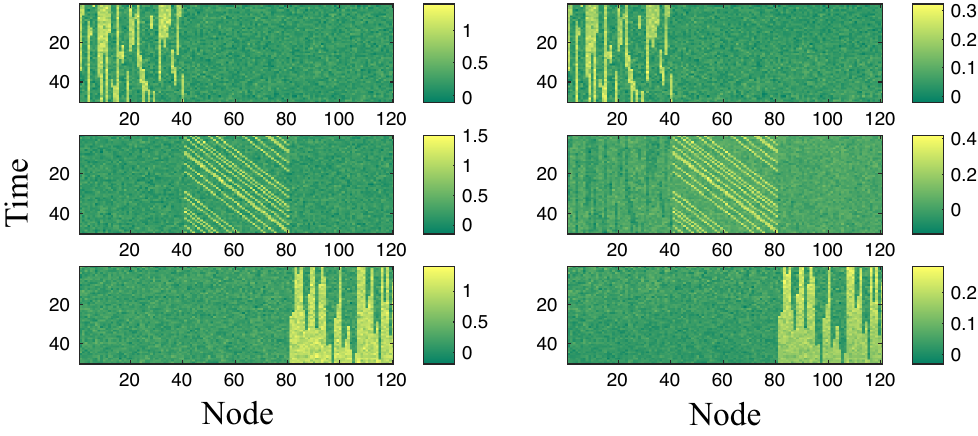}
\vspace*{-6mm}
\caption{\small{Exp.~3: Ground-truth evolving networks ${\bf B}_k$s (left) and recovered ones (right) ($\textrm{Noise}=1, \textrm{ridge},\textrm{FMS}_{\mB}=0.973$).}}
\label{fig:syn2B}
\end{figure}

\begin{figure}[t]
\centering
\includegraphics[width=\columnwidth]{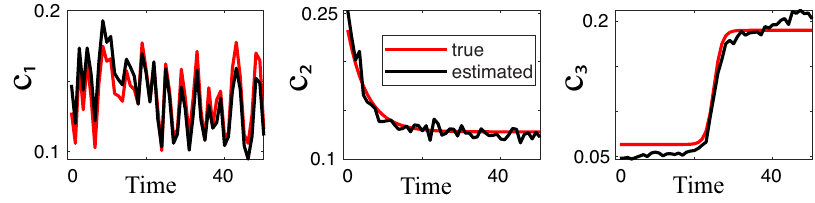}
\caption{\small{Exp.~3: Ground-truth temporal patterns $\bf C$ (red) and recovered ones (black) ($\textrm{Noise}=1, \textrm{ridge}, \textrm{FMS}_{\mC}=0.997$).}}
\label{fig:syn2C}
\end{figure}

\begin{figure}[t]
\centering
\includegraphics[width=0.9\columnwidth]{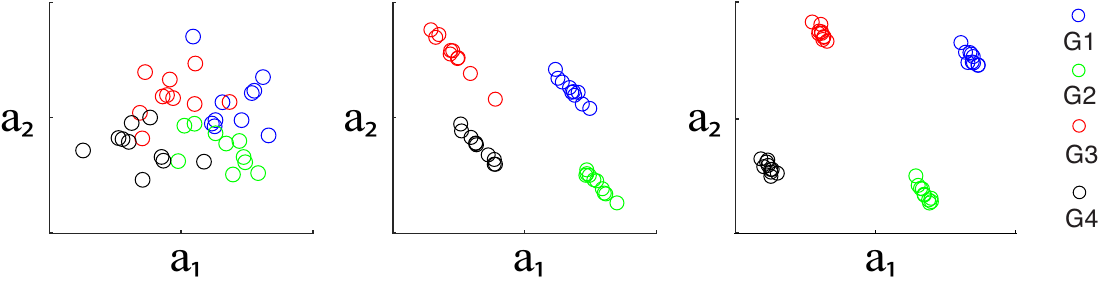}
\caption{\small{Exp.~3: Example of clustering structure in the ground-truth $\bf A$ (left: Noise=1) and recovered ones (middle: with coupling; right: with coupling and ridge regularization).}} 
\label{fig:syn2A_sm}
\label{fig:syn2A}
\end{figure}

In the data generation process, ${\mB}_k$s do not adhere to the PARAFAC2 constraint, but are designed to simulate evolving networks along the temporal mode $\mC$, similar to the experiment in \cite{RoBhJi20}. The factor matrices ${\mB}_k$ consist of three columns representing a shrinking, shifting, and growing network, respectively, and some background Gaussian noise as depicted in Figure~\ref{fig:syn2B}. $\mC$ represents a temporal pattern matrix encompassing an exponential, a sigmoidal, and a random curve, see Figure~\ref{fig:syn2C}). For the details of the data generation process, we refer to \cite{RoBhJi20}   \footnote{\url{https://github.com/marieroald/ICASSP20}}. Finally, a clustering structure with four clusters is incorporated in the first two columns of the coupled matrices $\mA$ and $\mE$ (Figure~\ref{fig:syn2A}) and $\mF$ is generated as a random non-negative matrix.
 Dataset sizes are $40 \times 120 \times 50$ and $40 \times 60$, and $R=3$ components are used. In this experiment, we do not add any noise to the datasets. Instead, we perturb matrix $\mA$ with Gaussian noise before constructing $\tX$. 
The performance of the model with exact coupling between $\mA$ and $\mE$ is compared to the one without any coupling. We impose non-negativity constraints on both $\mC$ and $\mF$. Table \ref{tab:exp2} shows the average performance in terms of FMS, model fit and clustering accuracy based on $k$-means clustering. Model fit and $\textrm{FMS}_{\mB}$ are never perfect for PARAFAC2 even in the noise-free and coupling-free case 
since true $\mB_k$s do not follow the PARAFAC2 constraint. Due to the fact that $\mA$ and $\mE$ are not equal, the model fit is better in the uncoupled case. However, the clustering performance increases through coupling and the coupled model still accurately captures the underlying patterns including the evolving networks which slightly violate the PARAFAC2 structure, as also depicted in Figures~\ref{fig:syn2B} and \ref{fig:syn2C}.
Adding ridge regularization with penalty parameter $10^{-4}$ on all modes further improves the clustering performance in the noisy case, see Table~\ref{tab:exp2} and Figure~\ref{fig:syn2A}.\\

\begin{table*}[t] 
\centering
\caption{\small{Average performance for experiment 3. FMS of $\mA$ using true noisy $\mA$ (and using clean $\mA$ ($\mA=\mE$)).}}
\small
\label{tab:exp2}
\renewcommand{\arraystretch}{1.2}
\begin{tabular}{cccccccccccccc} %
\toprule 
\multirow{2}{*}{Ridge}&\multirow{2}{*}{Coupling}&\multirow{2}{*}{Noise}& \multicolumn{2}{c}{Fit (\%)}&&\multicolumn{5}{c}{FMS}&&\multicolumn{2}{c}{Clustering acc. (\%)}\\
\cmidrule{4-5}\cmidrule{7-11}\cmidrule{13-14}
&&&PAR2&Matrix&&A&B&C&E&F&&A&E\\  
\midrule %
\multirow{6}{*}{\text{no}}&\multirow{3}{*}{\text{no}}&0& 99.98& 100&&1 (1)&0.99& 1& -&-&&100&-\\   %
&&0.5&99.98&100&&1 (0.89)&0.99&1&-&-&&95.63&-\\
&&1&99.98&100&&1 (0.71)&0.99&1&-&-&&65.88&-\\
\cmidrule{2-14}
&\multirow{3}{*}{\text{yes}}&0& 99.75& 100&&1(1)&0.99& 0.99& 1&1&&100&100\\   %
&&0.5&84.51&99.98&&0.90(0.98)&0.99&1&0.98&0.99&&100&100\\
&&1&56.67&99.95&&0.72(0.96)&0.98&0.99&0.96&0.99&&99.63&99.63\\
\cmidrule{1-14}
\multirow{3}{*}{\text{yes}}&\multirow{3}{*}{\text{yes}}&0& 99.91& 100&&1(1)&0.99& 1& 1&1&&100&100\\   %
&&0.5&83.05&99.97&&0.90(0.99)&0.98&1&0.99&1&&100&100\\
&&1&57.53&99.95&&0.73(0.99)&0.96&1&0.99&1&&100&100\\
\bottomrule %
\end{tabular}
\end{table*}

\subsubsection{Experiment 4:  Partial coupling and smoothness}
\begin{figure}
\vspace*{-0.2cm}
\centering
\includegraphics[width=\columnwidth]{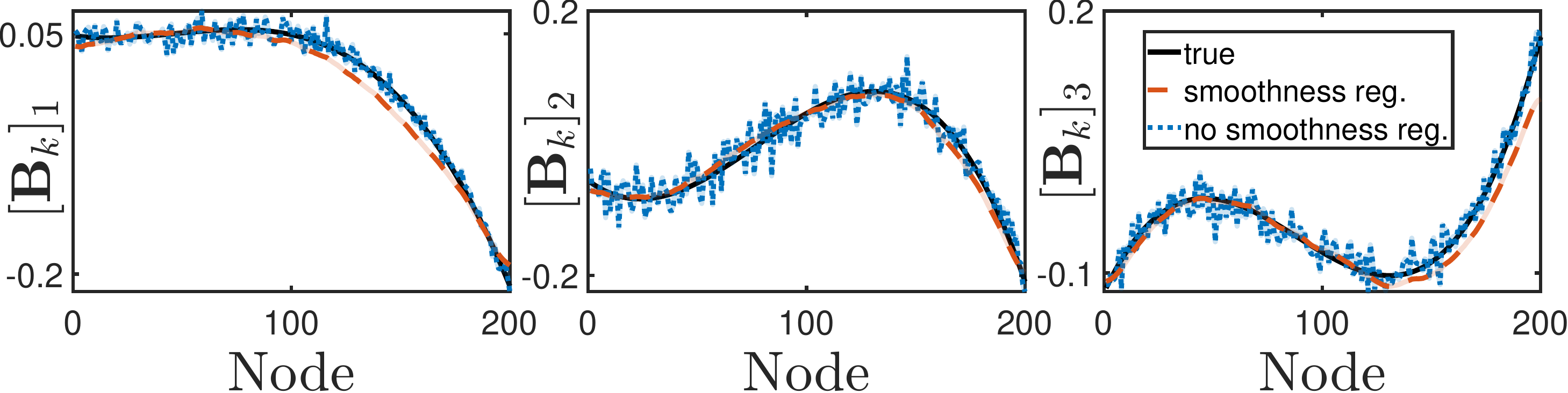}
\vspace*{-5mm}
\caption{\small{Exp.~4: Example of recovery of the three components of $\mB_k$, with and without smoothness regularization.}}
\label{fig:syn3A}
\vspace*{-5mm}
\end{figure}

Here, we show the versatility and utility of the proposed framework in a setting with partial coupling and smoothness regularization. We construct a tensor $\tX$ of size $30 \times\! 200 \times 30$ following a PARAFAC2 model, and a tensor $\tY$ of size $30 \times 20 \times 50$ following a CP model. Both have three components, but only two components are shared in the third mode (mode $\mC$ of PARAFAC2). The components in the $\mB_k$-mode are smooth, generated as in \cite{RoScCa22}. The coupled matrices $\mC$ and $\mE$ are non-negative and generated from a shifted $(+0.1)$ standard uniform distribution. All other factor matrices are drawn from the standard normal distribution. 
The noise level is set to $0.5$ for both datasets. 
We then solve the following optimization 
problem,
\begin{equation*}\small{
\begin{aligned}
    & \underset{\underset{
       \mA,\mE,\mF,\mG,\mDelta}{\mC,\left\lbrace \mB_k\right\rbrace_{k\leq K,}}}{\argmin} \frac{1}{2} \norm{\mY - \KOp{\mE, \mF,\mG}}_F^2  + \iota_{\mathcal{B}_{1,+}^2}(\mE)+ \iota_{\mathcal{B}_{1,+}^2}(\mC) \\
    &\  +\sum\limits_{k=1}^{K}\left[\frac{1}{2} \norm{\mX_k  - \mA \diag{\mC(k,:)} \mB_k^T}_F^2  + g_\texttt{GL}(\mB_k)\right]+ \iota_{\mathcal{B}_{1}^2}(\mA)\\
   & \ \ \ \  \text{s.t. } \ \ \ \ \ \ \ \ \ \ \   \left\lbrace\mB_k\right\rbrace_{k\leq K} \in \mathcal{P},\ \ \
   \mC =  \mDelta \hat{\mH}_{\mC}^{\Delta}, \mE = \mDelta  \hat{\mH}_{\mE}^{\Delta},
\end{aligned}}
\end{equation*}
where $g_\texttt{GL}$ denotes a columnwise graph laplacian regularization to promote smooth components, as in \cite{RoScCa22}. 
The factor vectors of modes $\mA,\mC$ and $\mE$ are constrained to be inside the unit $\ell 2$-ball $\mathcal{B}_1^2$ in order to make the smoothness regularization effective. Furthermore, modes $\mC$ and $\mE$ are constrained to be non-negative. We use linear coupling constraints 
to account for the partial coupling, \textit{i.e.,} $\hat{\mH}_{\mC}^{\Delta}$ 
indicates which columns from the ``dictionary" $\mDelta$ are present in $\mC$. The proposed algorithm recovers the true factors with an average FMS of $0.96$ for mode $\mB$, an average FMS of $0.97$ for mode $\mA$ and a FMS of $1$ for all other modes, 
yielding smooth $\mB_k$ components (Fig.~\ref{fig:syn3A}). A similar experiment with the same settings, but partial coupling between modes $\mA$ and $\mE$ can be found in the supplementary material. There, we obtain better accuracy in terms of the FMS. This is likely due to the fact that coupling with a CP model, which is more robust to noise than PARAFAC2, aids the accurate recovery of factor matrix $\mA$, which in turn stabilizes the estimation of the $\mB_k$ matrices in the presence of noise. Couplings in mode $\mC$ do not have the same effect, since the $\mB_k$ matrices are allowed to vary across mode $\mC$.

\subsection{Real Data}\label{sec:realdata}
\begin{figure*}[!t]
\vspace*{-0.3cm}
\centering
\includegraphics[width=2\columnwidth]{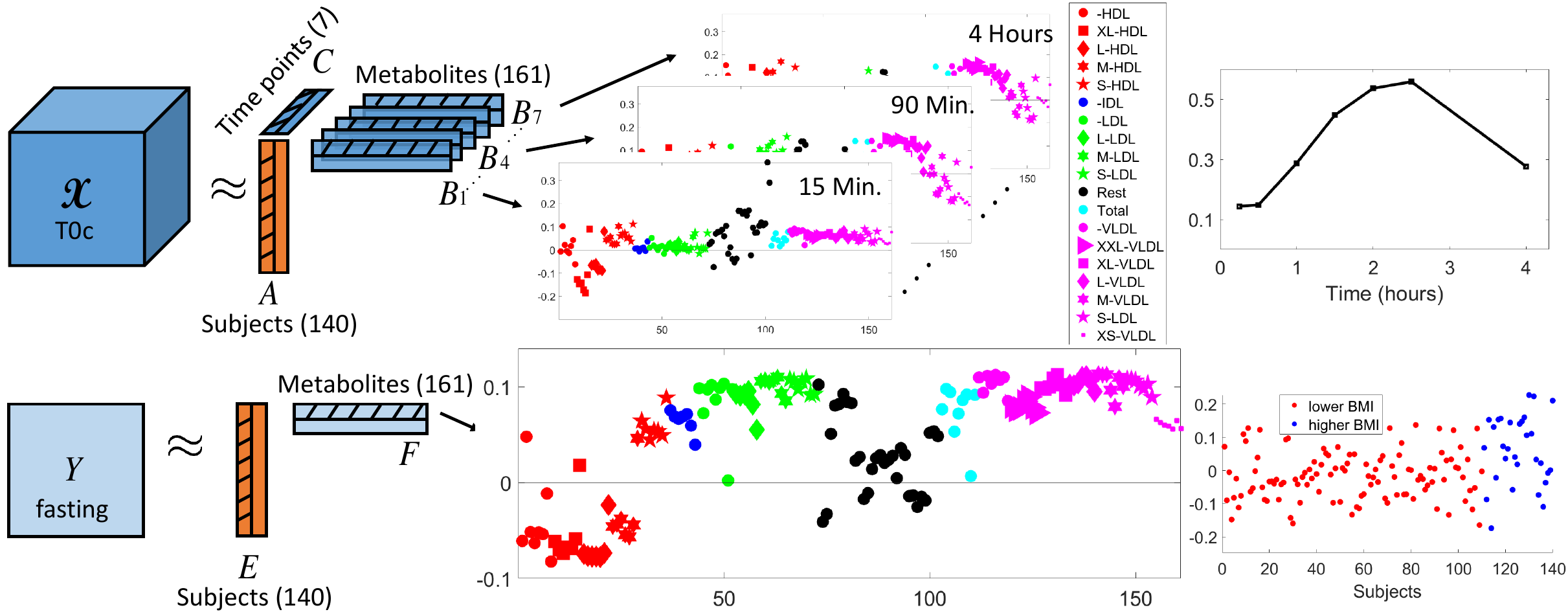}
\vspace*{-0.3cm}
\caption{Model for experiment \ref{sec:realdata} and extracted patterns in the first component.}
\label{fig:NMRmodel}
\vspace*{-0.3cm}
\end{figure*}

\begin{figure}[!t]
\vspace*{-0.3cm}
\centering
\includegraphics[width=0.8\columnwidth]{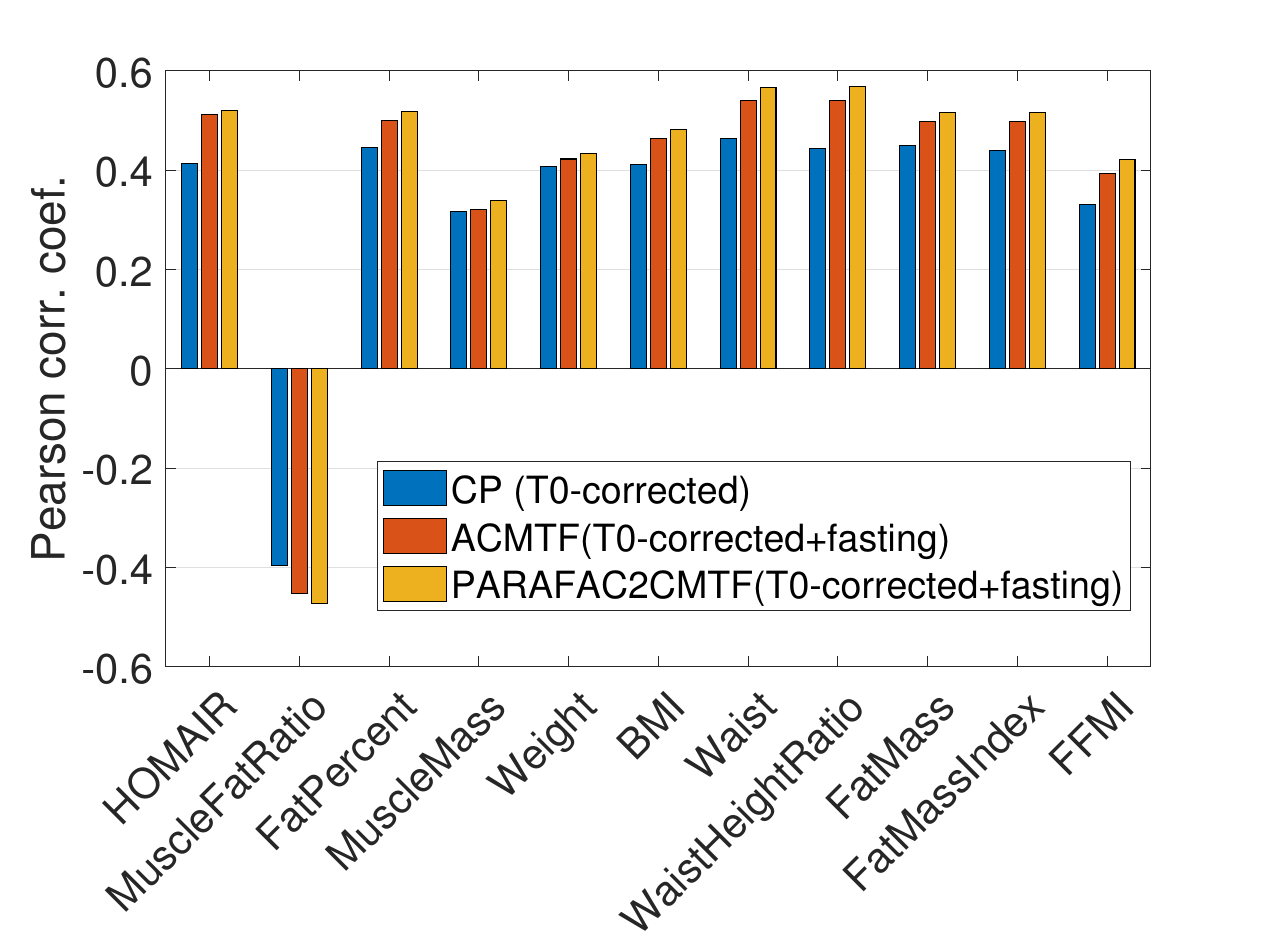}
\vspace*{-0.3cm}
\caption{Correlation between subject scores from  the component of interest and meta variables for the PARAFAC2-based CMTF model (PARAFAC2CMTF), CP-based CMTF model (ACMTF) \cite{LiYaHo24} and CP model of only the T0-corrected data \cite{YaLiHo24}. Descriptions of meta variables are as follows: HOMAIR: Homeostatic model assessment for Insulin Resistance; MuscleFatRatio: Muscle to fat ratio; FatPercent: Body fat percentage; MuscleMass: Amount of muscle in the body (kg); Weight (kg); BMI: Body Mass Index; Waist: Waist circumferance (cm); WaistHeightRatio: Waist measurement divided by height (cm); FatMass: Amount of body fat (kg); FatMassIndex; FFMI: Fat Free Mass Index.}
\label{fig:NMR_corr}
\vspace*{-0.5cm}
\end{figure}

In this experiment, we jointly analyze real dynamic and static (fasting state) metabolomics data from a meal challenge test from the $\text{COPSAC}_{2000}$ cohort (Copenhagen Prospective Studies on Asthma in Childhood) \cite{Bi04} to understand differences among subjects based on their metabolic response to a meal. Blood samples were taken after overnight fasting and at seven specific time points after a standardized meal. These samples were measured using Nuclear Magnetic Resonance (NMR) spectroscopy. The data includes both the NMR measurements and measurements of specific hormones. For more details about the data, see \cite{YaLiHo24,LiYaHo24}. The fasting state data is in the form of matrix $\mY$ with modes \emph{140 subjects} $\times$ \emph{161 metabolites}, and the dynamic data is in the form of a third-order tensor $\tX$ with modes \emph{140 subjects} $\times$ \emph{161 metabolites} $\times$ \emph{7 time points}. The tensor $\tX$ corresponds to T0-corrected data, where we correct for baseline differences by subtracting matrix $\mY$ from measurements at each time point. The data is further preprocessed as in \cite{LiYaHo24}.
A prior study has shown the importance of analyzing both fasting state
and T0-corrected data for understanding metabolic differences among subjects \cite{LiYaBa24}. Separate CP and PCA (principal component analysis) models on the T0-corrected and fasting state data, respectively, have revealed distinct dynamic and static metabolic patterns associated with BMI (body mass index)-related group differences  \cite{YaLiHo24}. Recently, joint \textit{static} and \textit{dynamic} biomarkers have been extracted from the same datasets using a CP-based CMTF model with coupling in the \textit{subjects} mode, demonstrating that the coupled model improves interpretability and strengthens potentially weak signals in separate datasets \cite{LiYaHo24}. However, the CP model of T0-corrected tensor extracts a fixed metabolic pattern for each component, which only changes in strength over time. Here, we propose to use PARAFAC2 to model the T0-corrected tensor with the goal of revealing evolving metabolite patterns in time.
We fit the coupled model \eqref{eq:costfuncA} to these datasets as in Figure~\ref{fig:NMRmodel}, where we use exact coupling in the subjects mode, \textit{i.e.} $\mA = \mDelta = \mE$, equal weights $w_{1,2}=0.5$, non-negativity constraints on the time mode $\mC$ and ridge regularization with penalty parameter $0.001$ on all modes. We use $R=2$ components based on the replicability of the extracted patterns (see supplementary material). The model fit is $53\%$ for $\mY$ and $31\%$ for $\tX$. The first component of this model is associated to a BMI-related group difference and shown in Figure~\ref{fig:NMRmodel}. The subject pattern $\mA(:,1)=\mE(:,1)$ in the lower right corner captures a statistically significant group difference in terms of BMI. Correlations of $\mA(:,1)$ with other meta variables are given in Figure~\ref{fig:NMR_corr} showing that this component is not only related to BMI but also to a phenotype defined by closely related variables. We observe that modelling of evolving metabolite patterns improves the correlations slightly while also revealing more insights about the underlying patterns. The corresponding static metabolic pattern $\mF(:,1)$ (middle, bottom of Figure~\ref{fig:NMRmodel}) shows the same biomarkers reported in \cite{LiYaHo24}. 
The pattern $\mC(:,1)$ extracted from the time mode (upper right corner of Figure~\ref{fig:NMRmodel}) exhibits again the peak around $2.5$ hours after the meal.
In contrast to \cite{LiYaHo24}, the dynamic metabolic pattern here is different for every time point. A detailed animation of this pattern can be found here\footnote{\url{https://github.com/AOADMM-DataFusionFramework/Supplementary-Materials/blob/master/evolving_metabolites.gif}}. At early time points (15 min and 30 min after meal intake), we see that changes in specific aminoacids, glycolysis related metabolites, insulin and c-peptide are positively related to high BMI. Over time, the metabolite pattern is increasingly dominated by triglycerides and large triglyceride-containing lipoproteins. Such increases are in line with expectations from the meal challenge context. Additionally, there are notable elevations in GlycA levels, which further underscores the expected metabolic shifts occurring in response to the meal challenge. The patterns of the second component, which is not associated to any available meta data, can be found in the supplementary material.

\section{Conclusion}
In this paper, we have presented a flexible algorithmic framework for PARAFAC2-based CMTF models based on ADMM. The framework allows to impose various constraints and regularizations on all modes, including the varying mode of PARAFAC2, and linear couplings in all static modes of PARAFAC2 with either a matrix-, a CP- or another PARAFAC2 model. Numerical experiments indicate that the proposed algorithm yields at least comparable and in many cases better performance in terms of computation time and accuracy compared to specialized algorithms for non-negative PARAFAC2 decompositions coupled with non-negative matrix- or CP-decomposition in the $\mC$-mode while achieving a much lower PARAFAC2 coupling residual without the need for tuning hyper-parameters. 

A limitation of the framework is that, currently, only Frobenius norm loss is supported for PARAFAC2 models by
the AO-ADMM framework. However, the splitting scheme used for the PARAFAC2 constraint allows
for a straightforward extension to other loss functions as proposed in our previous work \cite{ScCoAc21} for CP-based CMTF models. There, we use limited memory BFGS with bound constraints (L-BFGS-B) for the updates of the primal variables,  \textit{i.e.} the factor matrices $\mA,\mB_k,\mC,...$, in the case of non Frobenius norm, but differentiable loss functions like Kullback-Leibler and Itakura-Saito divergence or Huber loss. Another possibility is to extend the approach proposed in \cite{HuSiLi16} to PARAFAC2 models. This approach uses ADMM with another split variable in order to deal with different loss functions. It is therefore more efficient and can accommodate most common loss functions. However, we have shown in \cite{ScCoAc21} that it achieves less accurate results than using L-BFGS-B. We plan to extend the framework to different loss functions and datasets with missing entries. 

The proposed framework holds the promise to enhance knowledge discovery in many applications. In this paper, we have used a PARAFAC2-based CMTF model to jointly analyze static and dynamic metabolomics datasets coupled in the subjects mode and extract static and dynamic (evolving) markers of a specific phenotype. This corresponds to the case we have discussed under PARAFAC2-based CMTF models coupled in mode $\mA$. Another important application of PARAFAC2-based CMTF models is omics data fusion applications, where PARAFAC2 can account for variability across subjects by extracting subject-specific temporal patterns from longitudinal (metabolomics) data while jointly analyzing with other datasets. This would correspond to PARAFAC2-based CMTF models coupled in mode $\mC$. These applications are not limited to omics data analysis. For instance, TASTE has previously been used to extract such subject-specific temporal patterns in temporal phenotyping using EHR data \cite{AfPePa20} jointly analyzed with meta variables. Similarly, accounting for spatial and temporal variability across subjects and multimodal data fusion is an active research area in functional neuromaging data analysis, which can significantly benefit from the flexible modelling framework presented in this paper.


\section*{Acknowledgments}
This work was supported by the Research Council of Norway through project 300489. The authors would like to thank COPSAC for sharing the meal challenge data, and Parvaneh Ebrahimi, Age Smilde and Rasmus Bro for their valuable inputs on NMR data analysis and the interpretation of the results. Finally, we would like to thank Christos Chatzis for providing the animation of the evolving metabolite factor. The $\text{COPSAC}_{2000}$ study was conducted in accordance
with the Declaration of Helsinki and was approved by the Copenhagen Ethics Committee (KF 01-289/96 and H-16039498) and the Danish Data Protection Agency (2015-41-3696). Both parents gave written informed consent
before enrollment. At the $\text{COPSAC}_{2000}$ 18-year visit, when the blood samples were collected, the study participants gave written consent themselves.
\bibliographystyle{IEEEbib}
\bibliography{bibliography}

\begin{thebibliography}{10}

\bibitem{LaAdJu15}
D.~Lahat, T.~Adali, and C.~Jutten,
\newblock ``Multimodal data fusion: An overview of methods, challenges, and
  prospects,''
\newblock {\em Proceedings of the IEEE}, vol. 103, pp. 1449--1477, 2015.

\bibitem{AcShLe19}
E.~Acar, C.~Schenker, Y.~Levin-Schwartz, V.~Calhoun, and Tulay Adali,
\newblock ``Unraveling diagnostic biomarkers of schizophrenia through
  structure-revealing fusion of multi-modal neuroimaging data,''
\newblock {\em Frontiers in Neuroscience}, vol. 13, no. 416, 2019.

\bibitem{PrMaEa17}
N.~D. Price, A.~T. Magis, J.~C. Earls, G.~Glusman, R.~Levy, et~al.,
\newblock ``A wellness study of 108 individuals using personal, dense, dynamic
  data clouds,''
\newblock {\em Nature Biotechnology}, vol. 35, pp. 747–756, 2017.

\bibitem{BeVaDr20}
S.~E. Berry, Ana~M. Valdes, D.~A. Drew, F.~Asnicar, M.~Mazidi, et~al.,
\newblock ``Human postprandial responses to food and potential for precision
  nutrition,''
\newblock {\em Nature Medicine}, vol. 26, no. 6, pp. 964--973, 2020.

\bibitem{AcBrSm15}
E.~Acar, R.~Bro, and A.~K. Smilde,
\newblock ``Data fusion in metabolomics using coupled matrix and tensor
  factorizations,''
\newblock {\em Proceedings of the IEEE}, vol. 103, pp. 1602--1620, 2015.

\bibitem{YaLiHo24}
S.~Yan, L.~Li, D.~Horner, P.~Ebrahimi, B.~Chawes, L.~O. Dragsted, M.~A.
  Rasmussen, A.~K. Smilde, and E.~Acar,
\newblock ``Characterizing human postprandial metabolic response using multiway
  data analysis,''
\newblock {\em Metabolomics}, vol. 20, no. 50, pp. 1--14, 2024.

\bibitem{ErAcCe13}
B.~Ermis, E.~Acar, and A.~T. Cemgil,
\newblock ``Link prediction in heterogeneous data via generalized coupled
  tensor factorization,''
\newblock {\em Data Mining and Knowledge Discovery}, vol. 29, pp. 203--236,
  2015.

\bibitem{LiSuCaKo09}
Y.-R. Lin, J.~Sun, P.~Castro, R.~Konuru, H.~Sundaram, and A.~Kelliher,
\newblock ``{MetaFac}: Community discovery via relational hypergraph
  factorization,''
\newblock in {\em KDD'09: Proceedings of the 15th ACM SIGKDD Int. Conf. on
  Knowledge Discovery and Data Mining}, 2009, pp. 527--536.

\bibitem{PoSaVaAl11}
S.~P. Ponnapalli, M.~A. Saunders, C.~F.~Van Loan, and O.~Alter,
\newblock ``A higher-order generalized singular value decomposition for
  comparison of global {mRNA} expression from multiple organisms,''
\newblock {\em PLoS One}, vol. 6, pp. e28072, 2011.

\bibitem{Ba08}
L.~Badea,
\newblock ``Extracting gene expression profiles common to colon and pancreatic
  adenocarcinoma using simultaneous nonnegative matrix factorization,''
\newblock in {\em Pacific Symp. on Biocomputing}, 2008, pp. 279--290.

\bibitem{chatzichristos2022coupled}
C.~Chatzichristos, S.~{Van Eyndhoven}, E.~Kofidis, and S.~{Van Huffel},
\newblock ``Chapter 10 - {C}oupled tensor decompositions for data fusion,''
\newblock in {\em Tensors for Data Processing}, Yipeng Liu, Ed., pp. 341--370.
  Academic Press, 2022.

\bibitem{AcGuRa12}
E.~Acar, G.~Gurdeniz, M.~A. Rasmussen, D.~Rago, L.~O. Dragsted, and R.~Bro,
\newblock ``Coupled matrix factorization with sparse factors to identify
  potential biomarkers in metabolomics,''
\newblock {\em International Journal of Knowledge Discovery in Bioinformatics},
  vol. 3, no. 3, pp. 22--43, 2012.

\bibitem{LiYaHo24}
L.~Li, S.~Yan, D.~Horner, M.~A. Rasmussen, A.~K. Smilde, and E.~Acar,
\newblock ``Revealing static and dynamic biomarkers from postprandial
  metabolomics data through coupled matrix and tensor factorizations,''
\newblock {\em Metabolomics}, vol. 20, pp. 86, 2024.

\bibitem{ChDaEsKoTh18}
C.~Chatzichristos, M.~Davies, J.~Escudero, E.~Kofidis, and S.~Theodoridis,
\newblock ``Fusion of {EEG} and {fMRI} via soft coupled tensor
  decompositions,''
\newblock in {\em EUSIPCO: Proc. 26th European Signal Processing Conf.}, 2018,
  pp. 56--60.

\bibitem{RiDuGu15}
B.~Rivet, M.~Duda, A.~Guerin-Dugue, C.~Jutten, and P.~Comon,
\newblock ``Multimodal approach to estimate the ocular movements during {EEG}
  recordings: a coupled tensor factorization method,''
\newblock in {\em EMBC: Proc. 37th Annual Int. Conf. IEEE Engineering in
  Medicine and Biology Society}, 2015.

\bibitem{AcPaGu14}
E.~Acar, E.~E. Papalexakis, G.~Gurdeniz, M.~A. Rasmussen, A.~J. Lawaetz,
  M.~Nilsson, and R.~Bro,
\newblock ``Structure-revealing data fusion,''
\newblock {\em BMC Bioinformatics}, vol. 15, pp. 239, 2014.

\bibitem{FaCoCo16}
R.~C. Farias, J.~E. Cohen, and P.~Comon,
\newblock ``Exploring multimodal data fusion through joint decompositions with
  flexible couplings,''
\newblock {\em IEEE Transactions on Signal Processing}, vol. 64, no. 18, pp.
  4830--4844, 2016.

\bibitem{AlKaSi20}
F.~M. Almutairi, C.~I. Kanatsoulis, and N.~D. Sidiropoulos,
\newblock ``Tendi: Tensor disaggregation from multiple coarse views,''
\newblock in {\em PAKDD: Proc. 24th Pacific-Asia Conf. Advances in Knowledge
  Discovery and Data Mining}, 2020, p. 867–880.

\bibitem{KaFuSi18}
C.~I. Kanatsoulis, X.~Fu, N.~D. Sidiropoulos, and W.-K. Ma,
\newblock ``Hyperspectral super-resolution: A coupled tensor factorization
  approach,''
\newblock {\em IEEE Transactions on Signal Processing}, vol. 66, no. 24, pp.
  6503--6517, 2018.

\bibitem{EyHuDe17}
S.~Van Eyndhoven, B.~Hunyadi, L.~{De Lathauwer}, and S.~{Van Huffel},
\newblock ``Flexible data fusion of {EEG-fMRI}: Revealing neural-hemodynamic
  coupling through structured matrix-tensor factorization,''
\newblock in {\em EUSIPCO: Proc. 25th European Signal Processing Conf.}, 2017,
  pp. 26--30.

\bibitem{ApCuAg18}
A.~P. Appel, R.~L.~F. Cunha, C.~C. Aggarwal, and M.~M. Terakado,
\newblock ``Temporally evolving community detection and prediction in
  content-centric networks,''
\newblock in {\em ECML-PKDD: Proc. European Conf. Machine Learning and
  Principles and Practice of Knowledge Discovery in Databases}, 2018, pp.
  3--18.

\bibitem{Ha70}
R.~A. Harshman,
\newblock ``Foundations of the {PARAFAC} procedure: Models and conditions for
  an ``explanatory" multi-modal factor analysis,''
\newblock {\em UCLA Working Papers in Phonetics}, vol. 16, pp. 1--84, 1970.

\bibitem{CaCh70}
J.~D. Carroll and J.~J. Chang,
\newblock ``Analysis of individual differences in multidimensional scaling via
  an {N}-way generalization of ``{Eckart-Young}'' decomposition,''
\newblock {\em Psychometrika}, vol. 35, pp. 283--319, 1970.

\bibitem{Hars1972b}
R.~A. Harshman,
\newblock ``{P}{A}{R}{A}{F}{A}{C}2: {M}athematical and technical notes,''
\newblock {\em UCLA Working Papers in Phonetics}, vol. 22, pp. 30--44, 1972.

\bibitem{KiTeBr99}
H.~A.~L. Kiers, Jos M.~F. {Ten Berge}, and R.~Bro,
\newblock ``{PARAFAC2}---{Part I. A} direct fitting algorithm for the
  {PARAFAC2} model,''
\newblock {\em Journal of Chemometrics}, vol. 13, no. 3-4, pp. 275--294, 1999.

\bibitem{MaCh17}
K.~H. Madsen, N.~W. Churchill, and M.~Mørup,
\newblock ``Quantifying functional connectivity in multi-subject {fMRI} data
  using component models: Quantifying functional connectivity,''
\newblock {\em Human Brain Mapping}, vol. 38, no. 2, pp. 882--899, 2017.

\bibitem{BrAnKi99}
R.~Bro, C.~A. Andersson, and H.~A.~L. Kiers,
\newblock ``Parafac2-part ii. modeling chromatographic data with retention time
  shifts,''
\newblock {\em Journal of Chemometrics}, vol. 13, no. 3-4, pp. 295--309, 1999.

\bibitem{PePa19}
I.~Perros, E.~E. Papalexakis, R.~Vuduc, E.~Searles, and J.~Sun,
\newblock ``Temporal phenotyping of medically complex children via {PARAFAC2}
  tensor factorization,''
\newblock {\em Journal of Biomedical Informatics}, vol. 93, pp. 103125, 2019.

\bibitem{AfPePa18}
A.~Afshar, I.~Perros, E.~Papalexakis, E.~Searles, J.~Ho, and J.~Sun,
\newblock ``{COPA}: Constrained {PARAFAC2} for sparse \& large datasets,''
\newblock in {\em CIKM: Proc. 27th ACM Int. Conf. Information and Knowledge
  Management}, 2018, pp. 793--802.

\bibitem{RoBhJi20}
M.~Roald, S.~Bhinge, C.~Jia, V.~Calhoun, T.~Adalı, and E.~Acar,
\newblock ``Tracing network evolution using the {PARAFAC2} model,''
\newblock in {\em ICASSP: Proc. IEEE Int. Conf. Acoustics, Speech and Signal
  Processing}, 2020, pp. 1100--1104.

\bibitem{acar2022tracing}
E.~Acar, M.~Roald, K.~M. Hossain, V.~D. Calhoun, and T.~Adali,
\newblock ``Tracing evolving networks using tensor factorizations vs.
  {ICA}-based approaches,''
\newblock {\em Frontiers in Neuroscience}, vol. 16, pp. 861402, 2022.

\bibitem{ChNaHa18}
Y.~Cheng, K.~Naskovska, M.~Haardt, T.~Götz, and J.~Haueisen,
\newblock ``A new coupled {PARAFAC2} decomposition for joint processing of
  somatosensory evoked magnetic fields and somatosensory evoked electrical
  potentials,''
\newblock in {\em Proc. 52nd Asilomar Conf. Signals, Systems, and Computers},
  2018, pp. 806--810.

\bibitem{AfPePa20}
A.~Afshar, I.~Perros, H.~Park, C.~Defilippi, X.~Yan, W.~Stewart, J.~Ho, and
  J.~Sun,
\newblock ``{TASTE}: Temporal and static tensor factorization for phenotyping
  electronic health records,''
\newblock in {\em Proc. ACM Conf. Health, Inference, and Learning}, 2020, pp.
  193--203.

\bibitem{GuThPa20}
E.~Gujral, G.~Theocharous, and E.~E. Papalexakis,
\newblock ``{C3APTION}: Constraint coupled {CP} and {PARAFAC2} tensor
  decompostion,''
\newblock in {\em ASONAM: IEEE/ACM Int. Conf. Advances in Social Networks
  Analysis and Mining}, 2020, pp. 401--408.

\bibitem{SoHiMa23}
M.~D. {Sorochan Armstrong}, J.~L. Hinrich, A.~P. {de la Mata}, and J.~J.
  Harynuk,
\newblock ``Parafac2×n: Coupled decomposition of multi-modal data with drift
  in n modes,''
\newblock {\em Analytica Chimica Acta}, vol. 1249, pp. 340909, 2023.

\bibitem{CoBr18}
J.~E. Cohen and R.~Bro,
\newblock ``Nonnegative {PARAFAC2}: A flexible coupling approach,''
\newblock in {\em LVA/ICA}, 2018, pp. 89--98.

\bibitem{ScWaAc23}
C.~Schenker, X.~Wang, and E.~Acar,
\newblock ``{PARAFAC}2-based coupled matrix and tensor factorizations,''
\newblock in {\em ICASSP: Proc. IEEE Int. Conf. Acoustics, Speech and Signal
  Processing}, 2023, pp. 1--5.

\bibitem{KoBa09}
T.~G. Kolda and B.~W. Bader,
\newblock ``Tensor decompositions and applications,''
\newblock {\em SIAM Review}, vol. 51, no. 3, pp. 455--500, September 2009.

\bibitem{Hi27a}
F.~L. Hitchcock,
\newblock ``The expression of a tensor or a polyadic as a sum of products,''
\newblock {\em Journal of Mathematics and Physics}, vol. 6, no. 1, pp.
  164--189, 1927.

\bibitem{Kruskal77}
J.~B. Kruskal,
\newblock ``Three-way arrays: rank and uniqueness of trilinear decompositions,
  with application to arithmetic complexity and statistics,''
\newblock {\em Linear Algebra and its Applications}, vol. 18, no. 2, pp.
  95--138, 1977.

\bibitem{SiBr00}
N.~D. Sidiropoulos and R.~Bro,
\newblock ``On the uniqueness of multilinear decomposition of n-way arrays,''
\newblock {\em Journal of Chemometrics}, vol. 14, no. 3, pp. 229--239, 2000.

\bibitem{DoLa13b}
I.~Domanov and L.~De~Lathauwer,
\newblock ``On the uniqueness of the canonical polyadic decomposition of
  third-order tensors---part {II}: Uniqueness of the overall decomposition,''
\newblock {\em SIAM Journal on Matrix Analysis and Applications}, vol. 34, no.
  3, pp. 876--903, 2013.

\bibitem{EvLa22}
E.~Evert and L.~De~Lathauwer,
\newblock ``Guarantees for existence of a best canonical polyadic approximation
  of a noisy low-rank tensor,''
\newblock {\em SIAM Journal on Matrix Analysis and Applications}, vol. 43, no.
  1, pp. 328--369, 2022.

\bibitem{GoLo13}
G.~H. Golub and C.~F. Van~Loan,
\newblock {\em Matrix computations},
\newblock Johns Hopkins University Press, Baltimore, 4th edition, 2013.

\bibitem{HaLu96}
R.~A. Harshman and M.~E. Lundy,
\newblock ``Uniqueness proof for a family of models sharing features of
  {Tucker's} three-mode factor analysis and {PARAFAC/CANDECOMP},''
\newblock {\em Psychometrika}, vol. 61, no. 1, pp. 133--154, 1996.

\bibitem{BeJoKi96}
J.M.F. ten Berge and H.A.L Kiers,
\newblock ``Some uniqueness results for {PARAFAC2},''
\newblock {\em Psychometrika}, vol. 61, no. 1, pp. 123--132, 1996.

\bibitem{ScCoAc21}
C.~Schenker, J.~E. Cohen, and E.~Acar,
\newblock ``A flexible optimization framework for regularized matrix-tensor
  factorizations with linear couplings,''
\newblock {\em IEEE Journal of Selected Topics in Signal Processing}, vol. 15,
  no. 3, pp. 506--521, 2021.

\bibitem{RoScCa22}
M.~Roald, C.~Schenker, V.~D. Calhoun, T.~Adali, R.~Bro, J.~E. Cohen, and
  E.~Acar,
\newblock ``An {AO-ADMM} approach to constraining {PARAFAC2} on all modes,''
\newblock {\em SIAM Journal on Mathematics of Data Science}, vol. 4, no. 3, pp.
  1191--1222, 2022.

\bibitem{SoBaLa15}
L.~Sorber, M.~Van Barel, and L.~{De Lathauwer},
\newblock ``Structured data fusion,''
\newblock {\em IEEE Journal on Selected Topics in Signal Processing}, vol. 9,
  no. 4, pp. 586--600, 2015.

\bibitem{Vervliet2016Tensorlab}
N~Vervliet, O~Debals, L~Sorber, M~Van~Barel, and L~De~Lathauwer,
\newblock ``Tensorlab v3.0,''
\newblock {\em available online, URL: www.tensorlab.net}, 2016.

\bibitem{KiPa11}
J.~Kim and H.~Park,
\newblock ``Fast nonnegative matrix factorization: An active-set-like method
  and comparisons,''
\newblock {\em SIAM Journal on Scientific Computing}, vol. 33, no. 6, pp.
  3261--3281, 2011.

\bibitem{HyPa08}
H.~Kim and H.~Park,
\newblock ``Nonnegative matrix factorization based on alternating nonnegativity
  constrained least squares and active set method,''
\newblock {\em SIAM Journal on Matrix Analysis and Applications}, vol. 30, no.
  2, pp. 713--730, 2008.

\bibitem{KiHePa14}
J.~Kim, Y.~He, and H.~Park,
\newblock ``Algorithms for nonnegative matrix and tensor factorizations: a
  unified view based on block coordinate descent framework,''
\newblock {\em Journal of Global Optimization}, vol. 58, no. 2, pp. 285--319,
  2014.

\bibitem{Boyd2011Distributed}
S.~Boyd, N.~Parikh, E.~Chu, B.~Peleato, and J.~Eckstein,
\newblock ``Distributed optimization and statistical learning via the
  alternating direction method of multipliers,''
\newblock {\em Foundations and Trends in Machine Learning}, vol. 3, no. 1, pp.
  1--122, Jan 2011.

\bibitem{HuSiLi16}
K.~Huang, N.~D. Sidiropoulos, and A.~P. Liavas,
\newblock ``A flexible and efficient algorithmic framework for constrained
  matrix and tensor factorization,''
\newblock {\em IEEE Transactions on Signal Processing}, vol. 64, no. 19, pp.
  5052--5065, 2016.

\bibitem{Moreau1965Proximite}
J.-J. Moreau,
\newblock ``Proximit{\'e} et dualit{\'e} dans un espace hilbertien,''
\newblock {\em Bull. Soc. Math. France}, vol. 93, no. 2, pp. 273--299, 1965.

\bibitem{Be17}
A.~Beck,
\newblock {\em First-Order Methods in Optimization},
\newblock SIAM, 2017.

\bibitem{PaBo14}
N.~Parikh and S.~Boyd,
\newblock ``Proximal algorithms,''
\newblock {\em Foundations and Trends in Optimization}, vol. 1, no. 3, pp.
  127–239, jan 2014.

\bibitem{Tensortoolbox}
B.~W. Bader, T.~G. Kolda, and et~al.,
\newblock ``Tensor toolbox for {MATLAB}, version 3.5,''
\newblock {\em available online, URL: www.tensortoolbox.org}, 2023.

\bibitem{Bi04}
H.~Bisgaard,
\newblock ``The {Copenhagen} prospective study on asthma in childhood
  {(COPSAC)}: design, rationale, and baseline data from a longitudinal birth
  cohort study.,''
\newblock {\em Ann Allergy Asthma Immunol}, vol. 93, no. 4, pp. 381--389, 2004.

\bibitem{LiYaBa24}
L.~Li, S.~Yan, B.~Bakker, H.~Hoefsloot, B.~Chawes, D.~Horner, M.~A. Rasmussen,
  A.~K. Smilde, and E.~Acar,
\newblock ``Analyzing postprandial metabolomics data using multiway models: A
  simulation study,''
\newblock {\em BMC Bioinformatics}, vol. 25, 2024.

\end{thebibliography}


\vfill

\end{document}